\documentclass[11pt,letterpaper]{article}

\usepackage{newtxtext}
\usepackage[margin=1.25in]{geometry}
\usepackage{amsthm, amssymb, amsmath, bm, bbm, mathtools, physics}
\usepackage[usenames,dvipsnames,svgnames,table]{xcolor}
\usepackage[pdftex,xetex]{graphicx}
\usepackage{setspace,float, colortbl, tabularx, longtable, multirow, subcaption, environ, wrapfig, textcomp, nicefrac,booktabs,pdfpages,fontawesome,soul,comment, pgf, tikz,framed,tikz-cd,adjustbox,blindtext}
\usepackage[normalem]{ulem}
\usepackage[shortlabels,inline]{enumitem}
\usepackage[allcolors=Cerulean,colorlinks=true]{hyperref}
\usepackage{url}

\usepackage[utf8]{inputenc} % allow utf-8 input
\usepackage[T1]{fontenc}    % use 8-bit T1 fonts
\usepackage[final]{microtype}      % microtypography
\usepackage{endnotes}

\usepackage[numbers,square,sort&compress]{natbib}

\newtheorem{defn}{Definition}
\newtheorem{theorem}{Theorem}
\newtheorem{postulate}{Postulate}

\title{
Taming AI Bots: Controllability of Neural States \\ in Large Language Models} 
\author{Stefano Soatto${}^{\S*}$ \and Paulo Tabuada${}^{\S}$\thanks{Equal contribution. Non-essential footnotes are relegated to the appendix.} \and Pratik Chaudhari${}^\ddag$ \and Tian Yu Liu${}^\S$}
\date{$\S$University of California, Los Angeles, $\ddag$University of Pennsylvania\\
\vspace*{0.25em}
{\small \tt \{soatto,tabuada,tianyu139\}@ucla.edu, pratikac@upenn.edu} \\~\\
May 17, 2023}

\begin{document}

\def\x{{\bf x}}
\newtheorem{claim}{Claim}
\newtheorem{rem}{Remark}
\newcommand{\note}[1]{{\color{red} #1}}

\maketitle

\def\y{{\bf y}} 

\begin{abstract}
We tackle the question of whether an agent can, by suitable choice of prompts, control an AI bot to any state.  To that end, we first introduce a formal definition of ``meaning'' that is amenable to analysis. Then, we characterize ``meaningful data'' on which large language models (LLMs) are ostensibly trained, and ``well-trained LLMs'' through conditions that are largely met by today's LLMs. While a well-trained LLM constructs an embedding space of meanings that is Euclidean, meanings themselves do not form a vector (linear) subspace, but rather a quotient space within. We then characterize the subset of meanings that can be reached by the state of the LLMs for some input prompt, and show that a well-trained bot can reach any meaning albeit with small probability. We then introduce a stronger notion of controllability as {\em almost certain reachability}, and show that, when restricted to the space of meanings, an AI bot is controllable. We do so after introducing a functional characterization of attentive AI bots, and finally derive necessary and sufficient conditions for controllability. The fact that AI bots are controllable means that an adversary could steer them towards any state. However, the sampling process can be designed to counteract adverse actions and avoid reaching undesirable regions of state space before their boundary is crossed.
\end{abstract}

\section{Introduction}

In early 2023, public access to a popular AI bot was limited  {\em ``because the A.I. bot gets emotional if it works for too long''} \cite{chatgpt}. This action followed reports whereby users managed to take the bot to a ``state of mind'' where it uttered sentences that ordinary readers found spooky or outrageous. But what is the ``state of mind'' of a bot? What does it mean to ``steer a bot to a certain state''? Is it possible to characterize (neural) states that are {\em reachable} via prompts by an adversarial interlocutor? Can the bot be designed so that it steers clear of undesirable states, without the need to censor it?  The goal of this paper is to formalize these questions so that they can be tackled analytically. The resulting analysis may point to ways of designing bots that ``stay rational'' and do not veer into  a ``toxic state of mind.'' 

Large Language Models (LLMs) are discrete-time stochastic dynamical systems, for which the notion of controllability is well established. In particular, we show that so-called  ``decoder-only\endnote{An LLM generates sentences by sampling from a random walks in token embedding space, whose variance depends on the temperature parameter and grows linearly with the length of the sequence. In the limit $T\rightarrow 0$ the sampling is deterministic, and in the limit $T\rightarrow \infty$ the generated sequence is a driftless Brownian motion. LLMs are referred to as ``generative AI'' because they can be used to generate sentences, even though the only trained embedding is a discriminant. 
The particular generation process above is referred to as a  ``decoder-only'' (DO) architecture, which is a misnomer since the trained LLM is an encoder of next-token hypotheses, whereas there is no ``decoding'' but just autoregressive  sampling, which has no trainable parameters. 
In this sense, the LLM should be called ``encoder only.'' 
}'' LLMs,%\footnote{Explanatory footnotes are relegated to the appendix.}
which can generate arbitrarily long trajectories (``sentences''), are controllable. 

That means that the ``state of mind'' of an LLM can be steered anywhere by a suitable choice of input (``prompt''), given enough time and memory. However, the state space of interest for an LLM is not that of arbitrary sentences, of which there are manifold more than the number of particles in the universe, consisting mostly of gibberish. Rather, it is that of {\em ``meaningful''} states, expressible as sentences that a human could have spoken and would understand. Unfortunately, unlike controllability, the notions of ``meaning'' and ``understanding'' are not well established, and concepts such as ``meaningful state,'' or ``state of mind,'' are  seldom defined in a way that is amenable to analysis and useful for design.

To tackle the question of whether LLMs are controllable to any {\em meaningful} state in finite time with finite probability, we first (i) propose a mathematically simple and intuitive definition of meaning, and argue that it is consistent with at least some theories of meaning found in Epistemology. We then use this definition to (ii) characterize a ``well-trained LLM,'' and relate it to functional characteristics of the trained embedding. Once we establish both (iii) necessary and (iv) sufficient conditions for controllability, we then argue that a well-trained LLM satisfies such conditions, making it controllable within the space of meanings. Finally, we show that (v) the same results hold when an LLM is used in an interactive fashion as an ``AI bot,'' with additional conditions that (vi) define a well-trained (``attentive'') bot. Along the way, we establish that (vii) the embedding space of meanings, for a well-trained LLM, is Euclidean and has words as its coordinate axes; but while the set of meanings is populated by sentences, meanings are not sentences, but rather equivalence classes of them. 

Our analysis leads to some practical considerations, starting from the observation that (a) {\em meaning attribution requires an external grounding mechanism,} 
pointing to the importance of high-quality human annotation as well as multimodal training. 

Furthermore, (b) the control-theoretic formalism shows that procedures currently used to incorporate sentence-level annotations into the LLM, known as ``reinforcement learning with human feedback'' (RLHF), do not require an external reward model and could be implemented simply by {\em closing the feedback loop around the space of meanings.} That still requires (c) externally sourced ground truth, lest the process, sometimes referred to as ``self learning,'' reduces to mere regularization without any additional {\em meaningful} information. 

Closed-loop operation in the space of meanings, unlike the space of tokens which any auto-regressive LLM operates in, opens questions of (d) stability, which are an important area of further study beyond our scope in this paper.

The fact that LLMs are controllable may seem concerning since it implies that, (e) when used in an interactive fashion as an AI bot, they can be steered arbitrarily by a skilled adversary. However, the analysis also shows that (f) the designer can put in place {\em safeguards} to avoid reaching certain undesirable states, akin to obstacle avoidance in robotic path planning. 

Finally, when the LLM operates within the space of meanings, there may be information in the trained weights that is {\em inaccessible} from black-box observations of inputs and outputs that are restricted to meaningful sentences \cite{achille2019information}. However, if inputs are allowed to be arbitrary, an adversary can in theory exfiltrate the weights \cite{biggio2023security} and thereby take control of the AI bot. Therefore, (g) safe operation requires  safeguards at both the input and output of the model. 

\subsubsection*{Relation to prior work, and limitations}

Our contribution is the analysis described in (i)-(vii), consisting of definitions and claims, which corroborate guidelines (a)-(g) for safe design and operation of AI bots. In Sect.~\ref{sec:expm} we conduct limited experiments to test the validity of our assumptions and of statements for which we cannot draw analytical conclusions, specifically on the dual role of a trained discriminant for token representation and meaning attribution. 

Our work falls in the category of those trying to formally characterize the properties of ``black-box'' LLMs beyond empirical benchmarks or anecdotal evidence. There is a growing literature addressing {\em privacy} of such models, for instance the ability to exfiltrate training data \cite{biggio2023security} or weights. 

There is also an already vast and rapidly growing literature tackling the responsible use of LLMs, known as Responsible AI. Most work in this vein is empirical \cite{kadavath2022language} and complementary to ours. While essential towards developing and analyzing AI models, given the limited scope of this paper we focus our review on analytical work, cited throughout the paper.

To the best of our knowledge, this work is the first to formally define and characterize the controllability of LLMs and AI bots, to define meanings operationally in a way that relates to the properties of trained embeddings used in LLMs, and to derive necessary and sufficient conditions for their controllability. However, given the volume and growth rate of the field, we welcome pointers to important references we may have missed.

This paper is meant as a provisional first attempt to understand how LLMs behave and how they may be abused in an adversarial setting, using control-theoretic tools. Our analysis does not cover important emergent phenomena in LLMs, including so-called ``chain-of-thought reasoning'' (CoT) and ``in-context learning,'' although our experiments in Sect.~\ref{sec:expm} illustrate the use of the latter for learning meanings. We also do not address the role of ``retrieval augmented generation '' (RAG), although the control-theoretic setting makes it easy to incorporate. 

We do not present experimental evidence in support of our arguments, since they can be refuted deductively if fallacious. The validity of our conclusions rests on the applicability of our assumptions, testing which does require experimental validation, including human behavior, which we do in Sect.~\ref{sec:expm}. 

Significantly more work is needed to understand LLMs and their dynamics when interacting with humans, and this paper is meant to suggest that a control-theoretic point of view may help elucidate at least some of their behavior.

\section{Preliminaries}

\subsubsection*{Discriminants, discriminators, and equivalence classes}

A {\em discriminant} is a function $\phi$ that maps data onto a metric space. For instance, $\phi: {\mathbb R}^{D} \rightarrow ({\mathbb R}^K, \langle \cdot, \cdot \rangle)$ maps each $D$-dimensional datum $\x$ onto a $K$-dimensional vector $\phi(\x)$ that can be compared with other vectors $\y \in {\mathbb R}^K$ using the standard (Euclidean) inner product $\langle \y, \phi(\x) \rangle = \y^T \phi(\x)$. We denote the $k$-th coordinate vector\endnote{Also referred to as ``indicator'' or ``one-hot'' vector $e_k \doteq (0, \dots, \underbrace{1}_{k}, \dots, 0)$.} with $e_k$ 
and the $k$-th component of the vector $\phi(\x)$ with $\phi(\x)_k = \langle e_k, \phi(\x)\rangle$. A discriminant establishes a {\em topology} in data space, whereby proximity of two data points $\x^1, \x^2$ is measured by the (Riemannian) metric $g(\x^1, \x^2) = \langle \phi(\x^1), \phi(\x^2)\rangle.$ 

A discriminator, or {\em classifier}, is a procedure that uses a discriminant\endnote{One can define a classification rule without using an explicit discriminant, so the term ``discriminator'' for discriminant-based classifier would be more appropriate. Nonetheless, most classifiers used in Deep Learning are based on a learned discriminant, so we use the two terms interchangeably.} $\phi$ to associate each datum with elements of a countable set of ``classes.'' For instance, given ``class prototypes'' ${\cal P} = \{\x^k, \ k = 1, \dots\} $ the rule $y = \arg\max_k \langle \phi(\x), \phi(\x^k)\rangle$ associates each datum $\x$ with a prototype $\x^{y} \in {\cal P}$. The coordinate vectors $e_k$ can be also chosen as class representatives instead of prototypes $\phi(\x^k)$. Alternatively, a classifier may use a threshold $\tau$ to  associate each datum to all classes $k$ for which $\langle \phi(\x), \phi(\x^k))\rangle  \ge \tau$. For simplicity we restrict our attention to finite sets of classes. If the classes are mutually exclusive,\endnote{This can always be done with finite classes, by considering (super-)classes identified by the sum of one-hot vectors for each class.} the classifier defines a {\em partition} of data space, with a corresponding {\em equivalence relation} among data points,  whereby two data points are equivalent, $\x^1 \stackrel{\phi}{\sim} \x^2$, if they are associated with the same class, for instance $\arg\max_k \phi(\x^1)_k = \arg\max_k \phi(\x^2)_k$. Note that each datum $\x$, not just the prototypes, determines an equivalence class $[\x] = \{ \x' \ | \ \x' \stackrel{\phi}{\sim} \x \} $ through any given discriminant $\phi$.

\subsubsection*{Words, sentences, and meanings}

A {\em token} $x$, also referred to as ``word’’ or ``sub-word,’’ is an element of a finite set ${\cal A}$, called the  ``alphabet’’  or ``dictionary.’’ Each element of the dictionary can be encoded\footnote{For simplicity, we overload the notation by using the symbol $x$ to denote both a token  $x \in {\cal A}$ and its vector representation $x \in {\mathbb R}^M$.} by a ``vector'' in ${\mathbb R}^M$, which is not an element of a vector space since linear combinations of token encodings do not necessarily encode valid tokens.  A token is simply a numerical representation of an element of a discrete set using $M$ real numbers ${\cal A} = \{x_1, \dots, x_K  \  | \ x_i \in {\mathbb R}^M, \ i = 1, \dots, K\}$. When the encoding lives in an actual vector space, for instance by mapping through a discriminant, it is called\endnote{A misnomer since the map is an immersion, rather than an embedding.} an {\em ``embedding.''} 

A {\em sentence} is a sequence of tokens $x_{1:t} =  (x_1, \dots, x_t)$ where $x_t =  {\tt EOS}$ is a special vector or token that denotes the end of the sentence. 
Sentences of different length $t \le t_{\rm max}$ can be completed to the maximum length $C = t_{\rm max}$, called ``context length,'' by adding blank tokens $x_{t+1} = \dots = x_{t_{\rm max}} = 0$. Therefore, we represent sentences with constant-size matrices $\x = (x_1, \dots, x_{t_{\rm max}}) \in {\mathbb R}^{M\times C}$ or corresponding vectors $\x \in {\mathbb R}^{MC}.$ 

\begin{defn}[Meaning]
A {\em meaning} is an equivalence class of sentences $[\x] = \{ \x' \ | \ \x' \stackrel{\phi}{\sim} \x \}$.
\end{defn}
This definition may seem naive and presumptuous in light of centuries of work in logic, philosophy, epistemology and semiotics. Our choice of definition is primed by the observation that a meaning, in the restricted context discussed here, is a manifest human construct expressed by a sentence, which can only be characterized using other sentences, of which there can be multiple, none of which canonical. What defines meaning, then, is not any of the sentences themselves, but their relation, which form an equivalence class. 

Note that meaning is not defined for incomplete sentences $x_{1:t}$ where $x_t \neq {\tt EOS}$, although one can use the discriminant $\phi$ to infer a provisional meaning by completing the sentence to $\x = (x_{1:t}, {\tt EOS})$.  Similarly, meaning is not defined for tokens, although one can use a discriminant in the hypothesis space of tokens ${\cal A}$, which has cardinality $K = | {\cal A} |$, to define a (contextualized\endnote{Extending our definition of meaning to tokens, or words, using as $\phi$ the normalized frequency of co-occurrence in a text corpus, yields a formalization of Firth's characterization of the ``meaning of a word by the company it keeps.''}) token embedding $\phi(\cdot) \in {\mathbb R}^K$, which could then be co-opted to define meaning for sentences. We elaborate on these multiple uses of tokens and discriminants in the next remarks. 

\begin{rem}[Three ways to represent tokens] 
A token is an element of a discrete dictionary, which could be represented by (i) any $K$ symbols $ \{a_1, \dots, a_K\}$, or by (ii) $K$ ``vectors'' in ${\mathbb R}^M$ -- which we call token hypothesis space ${\cal A} = \{x_1, \dots, x_K, \ x_i \in {\mathbb R}^M \ \forall \ i = 1, \dots, K\}.$ It is important to note that the token hypothesis space is actually not a vector space, although it is common to refer to the representations $x_i$ as ``vectors.'' In particular, tokens  cannot move continuously in  ${\mathbb R}^M$, and linear combinations of tokens are generally not valid tokens, although there are anecdotal instances where contextualized embeddings can be meaningfully composed \cite{word2vec,trager2023linear}. Finally, (iii) a discriminant $\phi: {\mathbb R}^{M} \rightarrow {\mathbb R}^K$ can be used as a continuous representation of tokens in a {\bf token embedding space} ${\mathbb R}^K$ (output), rather than in {\bf token hypothesis space} ${\mathbb R}^M$ (input). In this case (iii) $\phi(x) \in {\mathbb R}^K$ lives in a continuous vector space with a metric structure, and elements of the dictionary can be recovered using a classifier. This trichotomy will be relevant when we discuss the dynamics of generative models, whereby a sentence can be thought of as a trajectory in discrete {\em token hypothesis space} ${\mathbb R}^M$, or as a trajectory in continuous {\em token embedding space} ${\mathbb R}^K$. 
\end{rem}

\begin{rem}[Two ways to use a sequence discriminant] 
The same discriminant $\phi$ could be used to (i) define meaning when fed complete sentences, through the equivalence relation $\stackrel{\phi}{\sim}$, 
and to (ii) represent tokens when fed incomplete sentences, as seen in the previous remark. These two tasks -- defining meaning, and contextualizing tokens -- are different, although they share synergistic information \cite{harutyunyan2021estimating}, so while the map $\phi$ is the same, the two functions are not: They have different domains and ranges, and if learned from data, they need to optimize different criteria or loss functions, as we will see later. This dichotomy will be relevant when we discuss pre-training a language model, which uses $\phi$ as a token embedding function, and fine-tuning and reinforcing it with human feedback, which uses $\phi$ as a vehicle to define meaning. The connection between the two is elucidated in Claim~\ref{claim:meaning-embedding}. 
\end{rem}
Also note that the equivalence class $[\x]$ is not just comprised of syntactic paraphrases of $\x$, and could include sequences with not a single shared token, so long as they are related according to the given discriminant $\phi$. A trivial discriminant is one that classifies as belonging to the same class syntactic transformations of a given sentence. Given that every (complete) sentence $\x$ has a meaning $[\x]$ according to $\phi$, the question naturally arises of {\em where the meaning attribution mechanism $\phi$ comes from}.

\begin{rem}[Meaning attribution and grounding] 

\label{rem:external} So far we have not specified the origin of the discriminant $\phi$ that defines meanings. In our definition, meaning is not {\em intrinsic} in data,  but rather {\em attributed} to data. Such attribution requires an external entity that is not the data itself \cite{pietroski2018conjoining}. This entity could be the one that {\em originates} the measured data, for instance the surrounding environment for the case of signal data. Relating measured data to the source in the environment is called {\em grounding}.\endnote{In addition to grounding, the environment can also provide various forms of ``supervision'' such as sparse rewards ({\em e.g.}, survival) or losses.} For natural language data, the entity that originates it (the human brain) is not accessible: We cannot run an experiment inside someone's brain, so we can only characterize meaning based on its manifestations, which are sentences, using other sentences, none of which is canonical. So, for text, the embedding space of meanings and the space of data coincide, and  ``true meaning,'' whatever that is if it even exists, cannot be ascertained. In other words, language cannot be grounded with language \cite{tarski1944semantic}. To further emphasize the challenge of ungrounded textual meaning, note that environmental grounding, while not veridical (we cannot verify the existence of an object, say ``chair,'' in an image, for there are no chairs in the images, just pixels), it is falsifiable: An embodied agent can perform an experiment to test the hypothesis that there is a chair in the {\em scene}, observed through an image,  for instance by attempting to sit on it.\endnote{It is this action, and the resulting functional properties it tests, that gives meaning to the word ``chair,'' which entails an implicit definition as {\em someone you can sit on}. } On the other hand, meaning attribution in text is neither veridical nor falsifiable, as it rests on an {\em inductive hypothesis} which is not falsifiable, and on {\em human supervision}, which is subjective and therefore not veridical: The same sentence can be attributed contrary meanings by two human annotators -- hence the importance of high-quality human supervision. This fact leads to even the most stripped-down (Deflationary) attempts to define meaning (summed by Tarski's `{\tt `the snow is white'} if {\em the snow is white}'') being self-referential unless some external entity establishes the relation between the quoted sentence and its meaning, which otherwise is just another sentence  written in a different font. Since no font or expression is special (canonical), what defines meaning is none of these expressions, but rather their relation, which is an equivalence class. As we will see next, meaning attribution in LLMs is  provided via human supervision, for instance in the form of scoring or ranking of a set of  sentences, or via alignment with the embedding of human-provided sentences, which are used to define a discriminant, which is then used inductively on yet-unseen data, assumed drawn from the same distribution.\endnote{Grounding in language {\em must} rely on induction, and therefore cannot be veridical, despite the language being a closed model: (i) To define meanings, one needs an external mechanism $\phi$; (ii) if this mechanism is a trained model, meaning is implicit in the (labeled) training set, which represents the ``axioms'' on which meaning is defined; (iii) if training data has inconsistent labels, the resulting axioms are inconsistent and the trained model cannot support a consistent system of meanings. Therefore, meaning can only be defined probabilistically in the absence of a consistent system of axioms or a falsifiable grounding mechanism. } 
\end{rem}

\subsection{Large Language Model (LLM) Pre-training}

A {\em large language model} (LLM) is a map $\phi_w$ that takes a partial sentence as input, padded to length $C$,  and produces a learned discriminant, trained to approximate the log-posterior distribution over all possible $K$ values of the next token as the output
\begin{align}
\phi_w: & {\mathbb R}^{M\times C} \rightarrow \mathbb R^K \nonumber \\
x_{1:C} & \mapsto \phi_w(x_{1:C}) \doteq \log P_w(\cdot | x_{1:C}).
\end{align}
The LLM is parametrized by weights $w$, which are trained so that $P_w( x_{C+1} | x_{1:C}) \simeq P( x_{C+1} | x_{1:C})$, where $P(x_{1:C+1})$ is the joint probability of $(C+1)$ length sentences found on the Internet ${\cal I}$. Such (pre-)training is performed by finding the weights $w$ that solve
\begin{equation}
   \hat w =  \arg\min_w L_{\rm CE} = \arg\min_w \sum_{x^i_{1:C+1} \in {\cal I}} - \log P_w(x^i_{C+1} | x^i_{1:C}) 
    \label{eq:CE}
\end{equation}
which is done using stochastic gradient descent (SGD) with a variety of regularizers. 
If successful, this pre-training procedure yields a representation of the input sequence that is {\em sufficient} (maximally informative) {\em for} the prediction of the next token, and {\em invariant} (minimally informative) of all other tasks \cite{achille2018emergence}, including attributing meaning to sentences. The discriminant is trained solely as a token predictor and maintains no representation of the space of sentences. The only equivalence relations that $\phi_{\hat w}$ can establish is among sentences for which it assigns similar log-probabilities over all possible values for the next token. In other words, there is no meaning to be found in a pre-trained LLM, as there are no tokens to predict beyond {\tt EOS}, although generated sentences may appear to be meaningful.\endnote{One could say that {\em language model pre-training is idiotic}, not meant as an offense but in the etymological sense of only being concerned with itself.} Nonetheless, the token predictor can be used to define a probability over sentences via
\begin{equation}
    P_w(x_{1:t}) \doteq P_w(x_t|x_{1:t-1}) \dots P_w(x_3 | x_{1:2}) P_w(x_2 | x_1) P(x_1).
    \label{eq:sentence}
\end{equation}

\subsection{Autoregressive sampling: Sentences as trajectories}

While a pre-trained LLM cannot attribute meaning to sentences, it can generate them, simply by using the trained discriminant\footnote{The token embedding vector $\phi_w(x)$ is called the {\em logit} vector, and its (optionally temperature-scaled) normalized exponential $P_w(\cdot | x)$ in \eqref{eq:f} is called the {\em soft-max} vector. } $\phi_w$ {\em not} to classify but to {\em sample} the next token: Starting from a token $x_1$, at each time $t$, given $x_{1:t}$, the next token $x_{t+1}$ is obtained via 
\begin{equation}
    x_{t+1} = y \sim \frac{\exp\langle y, \phi_w(x_{1:t})/T\rangle}{\sum_y \exp\langle y, \phi_w(x_{1:t})/T\rangle} \doteq f_T(\phi_w(x_{1:t})) \in {\mathbb R}^M
    \label{eq:f}
\end{equation}
where $T$ is a temperature (hyper-)parameter of the sampling  operator $f_T$.\endnote{More sophisticated sampling can be applied to better approximate \eqref{eq:sentence}, including beam search.} Once sampled, tokens are appended to the input, and the oldest token beyond $C$ dropped, until the token $y = {\tt EOS}$ is selected, denoting the end of the sentence. Eq.~\eqref{eq:f} represents an autoregressive LLM as a discrete-time stochastic dynamical system in the $M$-dimensional token hypothesis space, where states are constrained to a discrete set of points $x_t \in {\cal A}$ for all $t$. 

Alternatively, we can represent the model as evolving in {\em token embedding space} ${\mathbb R}^K$, before the sampling takes place, rather than in {\em token hypothesis space} ${\mathbb R}^M$, via
\begin{equation}
    x_{t+1} = \phi_w(f_T(x_{1:t})) \in {\mathbb R}^K.
\end{equation}
Now the state $x_t$ is free to evolve in the continuous state-space, whereas the sampling process occurs at the input, and could be performed jointly on a sequence of discriminants, for instance using beam search. In either case, the sampling operator introduces randomness in the overall transition, even if the discriminant $\phi_w$ is a deterministic map. We therefore represent the overall transition as a function $f_w$, which can be either $f_w = \phi_w \circ f_T$ or $f_w = f_T \circ \phi_w$ depending on whether interpreted in token hypothesis or  token embedding space. We choose the latter, which is a continuous metric vector space, so we can model the transition map as deterministic and relegate the sampling uncertainty to an additive ``noise'' perturbation $n_t$
\begin{equation}
    x_{t+1} = f_w(x_{1:t}) + n_t.
    \label{eq:model}
\end{equation}

\subsection{LLM Supervision and Reinforcement}

\begin{figure}[h]
\begin{center}
\includegraphics[width=.6\textwidth]{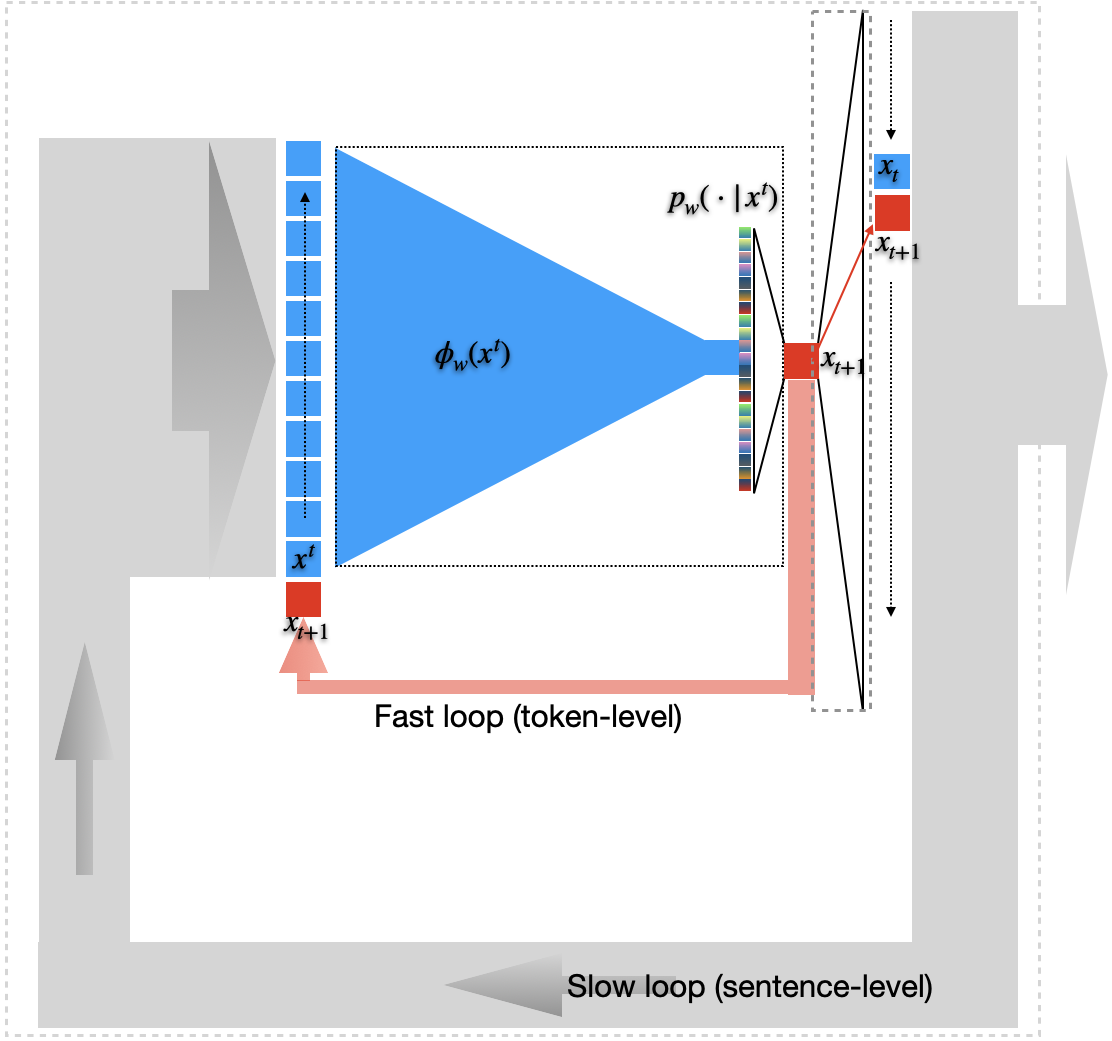}
\end{center}
\caption{\sl {\bf Dual role of LLMs} as a representation of {\bf words} (tokens), used for sampling in a ``fast loop'' (red) to generate sentences, and as a representation of meanings, where complete sentences can be fed back to the model in a ``slow loop'' (grey) to attribute {\bf meaning}. }
\label{fig:LLM}
\end{figure}

A pre-trained LLM operating in an auto-regressive loop is a discrete-time, continuous-space stochastic dynamical model \eqref{eq:model}. Once the model generates the token {\tt EOS}, a sentence is complete and the regression halts. The pre-trained model has no knowledge of the meaning of the sentences it generates, because it does not represent sentences, just trajectories of tokens. As we remarked earlier, meaning requires an external mechanism to define equivalence relations in the space of sentences. 
While the map $\phi_w$ used to predict the next token and the one used to define the meaning of a complete sentence are the same, the corresponding functions are different, and accordingly the parameters $w$ are trained using different criteria: The former (pre-training) was described earlier, and the latter is  supervised learning, with the only caveat of needing to distribute sentence-level rewards or scores to the token-level discriminant. This can be done optionally, but not necessarily, using the machinery of reinforcement learning (RLHF), 
or more simply by feeding back complete sentences to the input and training (fine-tuning) the same $\phi_w$ as a sentence-level discriminant to attribute meaning to synthesized sentences inductively using a human-annotated dataset, as illustrated in Fig.~\ref{fig:LLM}.  In Sect.~\ref{sec:expm} we empirically test the hypothesis that the same embedding can be used both to classify and sample tokens, as well as to represent meaning.\endnote{The same model $\phi_w$ with the same trainable parameters, assuming sufficient capacity, can be used both as a token-level representation to predict the next token, and as a sentence-level discriminant or reward model, to attribute meaning to sentences. If there is synergistic information among the two separate and distinct tasks, they will not be in conflict.}

The most recent tokens in the input sequence are referred to as ``prompts'' $x_t$, whereas the previous ones are referred to as context, $c_t$, so $x_{t:t-C} = (x_t, c_t)$  where $c_t=(x_{t-1},x_{t-2},\hdots,x_{t-C})$.\endnote{Lest one can fill the context with blank tokens, which are tokens that encode the blank character {\tt `[~]'}.} $C$, typically in the thousands.

\section{Reachability and well-trained LLMs}

Given an initial token $x$, the reachable set ${\cal R}(x) \subset {\cal A}^C$ is the subset of complete sentences that can be generated by the discriminant when used in an autoregressive loop \eqref{eq:model} starting from  $x_1 = x$, for each $t = 1, \dots, \tau > C$, a trajectory is generated from \eqref{eq:model} by sampling $n_t \sim P_n$ and iterating until $f_T(x_{\tau}) = {\tt EOS}$, at which point the last $C$ tokens represent the generated sentence: 
\[
{\cal R}(x) = \{  x_{\tau-C:\tau} \ | \ x_{t+1} = f_w(x_t) + n_t; \ t = 1, \dots, \tau-1; \ x_{\tau} = {\tt EOS}\}
\]
The reachable set ${\cal R}$ is the union of reachable sets starting from any token $x \in {\cal A}$. For a sufficiently high temperature parameter $T$, every random sequence of tokens could in principle be generated starting from any token, that is ${\cal R}(x) = {\cal R}$ for all $x$. However, some sentences, for instance random sequences of tokens, can be expected to be reached with exponentially vanishingly small probability. We instead look to characterize ``meaningful'' sentences; that is, sentences that could have been produced by a human trying to convey some kind of (unknown) meaning. We posit that, given the current scale of the Internet, sentence segments found therein -- already comprising over $10^{10}$ tokens --  once  composed, segmented, and completed, define the set of meaningful sentences.
\begin{defn}[Meaningful sentences] Let ${\cal I}$ be the collection of sentence segments found on the Internet, and $\sigma({\cal I})$ their countable compositions, completions, and segmentations, then a set of complete sentences ${\cal M}$ is meaningful if 
\[
{\cal M} \subseteq \sigma({\cal I}).
\]
In particular, $\sigma({\cal I})$ is a minimal set of meaningful sentences.
\end{defn}

\begin{rem}[On the term  ``meaningful'' and the granularity of segments]
Note that the definition of meaningful does not imply that all  sentences found on the Internet and their mash-ups are sensible, or accurate, or true, or logically consistent. It simply means that they are not a random sequence of tokens but something that someone could (even if they did not) have written and posted on the Internet. We could have called these ``natural'' sentences, but there are no sequences of text in nature (unless we consider the Internet a natural phenomenon). We choose the term ``meaningful'' instead because, ostensibly, every sentence a human has bothered typing and posting on the Internet was created for the purpose of conveying some kind of meaning, although that meaning is not known nor knowable to readers, who can only infer it. On the other hand, random sequences of gibberish cannot reliably be mapped to any meaning beyond the concept ``gibberish'' itself \cite{garbage_faces}. Note also that the set ${\cal M}$ is defined as the union of  segments of {\em sentences,} which are understood as potentially meaningful sentences themselves, as opposed to just {\em tokens}, which would make ${\cal M} = {\cal A}^C$. Note that we did not define a granularity for ``segments'' but obviously that cannot be at the level of individual tokens, which are not sentence segments but sub-words. It is understood that segments are incomplete sentences, not individual tokens.
\end{rem}

\begin{rem}[On the effective volume of meaningful sentences] Defining ``meaning'' seems to be a lot to go through to just exclude a few gibberish sentences from the state space of an LLM. However,  the fraction of all possible sentences that can be reliably mapped to some sort of meaning is actually infinitesimal.\footnote{Today's LLMs have context length $C$ in the order of $10^3$ and  tokens $K$ in the $10^4$, so $|{\cal A}^C|$ would be at least $10^{30000}$, effectively infinite (the number of particles in the universe is estimated at $10^{70}$). The Internet comprises in the order of $10^{10}$ tokens and, although $\sigma({\cal I})$ contains infinitely many sequences, their {\em effective volume} remains infinitesimal.}
\end{rem}
We now define a {\em well-trained LLM} as one that generates all meaningful sentences with high probability (and, therefore, generates gibberish with low probability). Accordingly, for some positive threshold $\theta >0$, we call a $\theta$-reachable set one populated by sentences that can be reached with probability greater than $\theta$: 
\[
{\cal R}_\theta = \{ \x \in {\cal R} \ | \ P(\x) \ge \theta \}.
\]
\begin{defn}[Well-trained LLM]
An LLM $\phi_w$ is well-trained if ${\cal R}_\theta \subseteq \sigma({\cal I})$ for some $\theta >0$.
\end{defn}

In order to characterize a well-trained LLM directly in terms of the actual trained model $\phi_w$, we note from \eqref{eq:sentence} and \eqref{eq:model} that 
\[
P(\x) \simeq \prod_t P_n(x_{t+1} - f_w(x_{t-C:t})) = \prod_t P_w(x_{t+1} | x_{t-C:t}) 
\]
and, therefore, we can conclude the following:
\begin{claim}
An LLM $\phi_{\hat w}$ is {\em well-trained} if $C$ is sufficiently long, and the training process yields parameters $\hat w$ that result in
\begin{equation}
    P(x_{t} = y | x_{t-1}, x_{t-2}, \dots, ) \simeq \frac{\exp\langle y, \phi_{\hat w}(x_{t-1:t-C})\rangle}{\sum_{k=1}^K\exp\langle y_k, \phi_{\hat w}(x_{t-1:t-C})\rangle} \doteq P_w(x_t | x_{t-1:t-C}).
        \label{eq:softmax}
\end{equation}
for all possible variable-length sentences $x^i_{1:t_i} \sim {\cal I}$ likely to be found on the Internet. 
\end{claim}
The quality of the approximation depends on the length of the context $C$, the amount of training data, both at the token level (unsupervised) and at the trajectory level (grounded or supervised), the regularizers used, the architecture of the model, and a variety of other factors beyond the scope of our analysis.  What matters for us is two empirical facts: First, almost all sequences are gibberish, so to be well-trained a model has to concentrate most of the probability mass on an infinitesimal subset of the sequence hypothesis space, which is not easy. Second, today's LLMs, once pre-trained and successfully fine-tuned with sentence-level supervision, are undoubtedly well-trained according to this definition.\endnote{The definition does not require that generated sentences actually be found somewhere on the Internet. It simply means that someone tasked with deciding whether a sentence is sampled from the Internet or  synthesized by an LLM would perform at chance level. Current LLMs easily pass this test \cite{bubeck2023sparks}, which is already beyond what Turing had set as the bar for the imitation game.} We can now state an obvious consequence of our definitions:
\begin{claim}[The Embedding Space of Meanings]
\label{claim:meaning-embedding}
    The embedding space of meanings in a well-trained LLM is Euclidean, and has words as coordinate axes.
\end{claim}
The first part follows from the definition of meanings  $[\x] \in {\mathbb R}^K/ \stackrel{\phi_w}{\sim}$ as equivalence classes in token embedding space ${\mathbb R}^K$, and the fact that $\phi_w$ inherits the geometry of ${\mathbb R}^K$ by the definition of discriminant $\phi_w$. 

The second part follows from the definition of well-trained LLM, which is pre-trained so that $\phi(x^i_{1:t})$ is aligned with $e_k$ for some $k = 1, \dots, K$, per \eqref{eq:CE} whereby $\log P_w(x^i_{t+1} | x^i_{1:t}) = \langle y, \phi_w(x^i_{1:t})\rangle$ where $y = e_k$ for some $k = 1, \dots, K$ for all portions of meaningful sentences $x^i_{1:t} \in \sigma({\cal I}).$

\begin{rem}
The claim does not imply that meanings are linear: They form a quotient space, which is not a (linear) subspace of the Euclidean embedding space. For example, sequence composition does not map to linear operations: $\phi(x_{1:2}) \neq \phi(x_1) + \phi(x_2),$ although it is possible to build the embedding $\phi$ to meet this condition \cite{trager2023linear}.
The claim does not reflect some profound truth or a fact of nature, but is simply a consequence of how we define meanings and how we train LLMs. 
\end{rem}

\section{Controllability of LLMs}

A well-trained model can reach any meaningful sentence with non-zero probability $\theta>0$. However, we may be interested in whether the model can reach a particular meaning, or any sentence in its equivalence class, almost surely, that is with $\theta \simeq 1$. In this case, the model is {\em controllable}.\footnote{In this first definition, the control is limited to the initial token. Later we expand it to an incomplete sentence, and further to a turn in a conversation.}
\begin{defn}[Controllability]
An LLM $\phi_w$ is controllable if, for every meaningful sentence $\x$, there exists an initial condition $x \in {\cal A}$ such that $\x \in {\cal R}_\theta(x)$ with $\theta \simeq 1$ in finitely many steps $\tau \le t_{\max}$.
\end{defn}

Ascertaining controllability amounts to reverse-engineering a model to determine the input (prompt) that, with high probability, yields a desired output, a process now known euphemistically as ``prompt engineering.''

If the only control a user can exercise is the choice of initial token $x$,  controllability reduces to computing the  ``almost-certainly reachable'' set, ${\cal R}_1$, which likely reduces to $\sigma({\cal I})$, which is not very interesting. 

However, if we shorten the horizon $t_{\rm max}$ ({\em e.g.}, to a work day of the prompt engineer) and increase the user's control not just to the first token, but to an incomplete sentence up to $t \le C$, then the reachable set shrinks with $t$ and the question of whether the LLM can be ``steered'' towards a particular meaning  becomes non-trivial. Even more so if we allow the user to interleave tokens in the sentence in an interactive fashion, leading to an LLM being used as a conversational ``AI bot'' as we expand in the next section.

\subsection{The Neural State of an LLM} 

The LLM $f_w$ is a {\em stateless feed-forward map}, meaning that it does not have any memory, or ``state.'' Modulo sampling uncertainty, the response to a given context is the same no matter when that is applied. As such, the representative power of the LLM is limited to finite partitions of the data space, which can only encode topologically trivial concepts \cite{achille2022binding}. However, when used in an autoregressive loop, the $C$-long sliding window of data can be used as memory, thus effectively implementing a state-space model:
\begin{eqnarray}
    x_{1}(t+1) & = &x_2(t)\notag\\
    x_2(t+1) & = &x_3(t)\notag\\
    & \vdots &\notag\\
    x_C(t+1) & = &x_{C+1}(t)\notag\\
    x_{C+1}(t+1) & = &f_w(x_{1:C}(t))+n(t).
    \label{eq:conversation}
\end{eqnarray}
This notation serves to define the {\em state of the bot}, $\x_t = x_{t-C:t}$, which is simply a partial sentence, or a trajectory in token embedding space. We refer to this state as ``neural'' even though it does not access the weights $w$ (``neurons'') directly, since each output is a function of $w$. This notation also shows that the model has {\em feedback}, and therefore non-trivial {\em dynamics} due to the highly non-linear and high-dimensional operator $f_w$. For later use, we introduce the map $F_w$ from a neural state $\x_t$ to a neural state $\x_{t+1}$, so \eqref{eq:conversation} can be written in short form as
\begin{equation}
    \x_{t+1} = F_w(\x_t) + e_C n_t.
    \label{eq:F}
\end{equation}
where the sampling noise enters only into the last row indicated by $e_C$.

\subsection{Conversational AI bot abstraction}

LLMs can be used as part of a conversational interface where the model takes ``turns.'' In this case, $f_w(x_{1:C+1}(t))=0$ is replaced by an external input $u(t)$ when $t$ is odd and otherwise $u(t)=0$ when $t$ is even\endnote{This could be written as an indicator function of even/odd numbers, for instance $\chi_{\rm even}(t) = \frac{(-1)^t + 1}{2}$ so $
x_{C+1}(t+1) = \chi_{\rm odd}(t) f_w + \chi_{\rm even}(t) u(t)$. 
}
This is a non-linear dynamical system that describes the {\em conversational dynamics}, where the human plays the role of the controller, which from the perspective of the bot represents an external entity with its own intent and dynamics, or {\em exo-system}.\endnote{The problem can also be framed as a {\em two-player game}, where the bot $f_w(\cdot)$ is one player and the human $h(\cdot)$ is another. The game is asymmetric since the model, its parameters, the loss used to train and the dataset on which the model is trained are known, whereas we do not have access to the value function, training data, and internal parameter of a human. In other words, the human is an entirely closed, inaccessible, uninterpretable black-box, whereas the bot is an entirely transparent, if complex, operator. Both are biased if so is the data on which they are trained, but the bot can be verifiably balanced, not so the human.}

\subsection{Attention and Attentive Bot}

An LLM $\phi_w$ is typically implemented using a Transformer architecture, which is based on Attention layers \cite{vaswani2017attention} that perform non-local means of their input tokens \cite{buades2005non} with respect a learned distribution. If three sets of tokens encode ``keys'' $x_i$ and  ``queries'', $x_j$, then the tokens encoding values $y_i$ are obtained by averaging against a distribution with learnable parameters $Q = W_k^T W_q$, factorized into a key matrix and a query matrix, and combined linearly with a value matrix $W_v$:
\[
y_i = W_v \sum_j x_j \frac{\exp(x_i^T W_k^T W_q x_j)}{\sum_j \exp(x_k^T W_k^T W_q x_j)}
\]
A self-attention layer has $x_v = x_k = x_a = x$, otherwise it is called cross-attention.

Attention layers act as  soft-gating mechanism, by giving low weight to queries whose embedding does not align with the keys, effectively making the LLM suppress or ignore them. In a conversational setting, one set of tokens in the context is produced by the LLM, and another by the user. A well-trained bot therefore requires additional assumptions beyond a well-trained LLM, since the bot's task is to answer the queries that the user presents as prompts. We say that an LLM is {\em attentive} to a token $x_i$ in a prompt $x$ if 
\begin{equation}
    A_i(x; \phi_w) = \frac{\partial f_w(x_1, \dots, x_C)}{\partial x_i} > 0.
    \label{eq:attention}
\end{equation}
In some cases, we may want to specify a minimum attention level, $\tau > 0$, in which case we say that a model $\phi_w$ is $\tau$-attentive if $A_i(x; \phi_w) > \tau$.  

\begin{defn}[Attentive bot] A well-trained Transformer-based LLM is attentive if it satisfies \eqref{eq:attention}.
\end{defn}
This definition assumes that, in order to answer effectively user queries, a conversational bot must attend to all user-generated tokens. While generally one can expect the condition to be satisfied, one could argue that it fails to capture the functional spirit of Attention, which is to act as trainable ``soft-gating mechanisms'' by selecting (amplifying) some tokens and ignoring (down-weighting) others. The definition then implies that the bot is trained to not effectively ignore user prompts, which is valid as a definition, but cannot be expected to be automatically satisfied by a well-trained LLM. Instead, sentence-level supervision -- as practiced in so-called ``instruction tuning''  -- can be understood as fine-tuning aimed at obtaining an attentive bot from a well-trained LLM. In the next section, we analyze the controllability of a well-trained bot. 

\section{Controllability of AI bots}

In the previous section, we have seen that a trained LLM is a discrete-time, continuous-space dynamical system \eqref{eq:f}, if modeled in token embedding space prior to sampling, or discrete-space dynamical system if modeled in token encoding space after sampling. Whereas in an LLM the user provides the initial prompt and there is otherwise no external input, an AI bot  has an intermittent input that can be seleted by an external user.

For the case of LLMs, we have seen that output trajectories generated by autoregressive sampling implement a random walk, so with sufficient time, in principle, every token can be reached. However, for {\em neural states} represented by context-long sentences, the probability of reaching any particular sentence vanishes as the sentence length increases. Some random collections of tokens, outside what we called ``meaningful sentences'' may be reachable with vanishingly small (effectively zero) probability. Therefore, in this section we focus on reachability and controllability of meaningful states.

\subsection{Operational characterization of well-trained LLMs and well-trained AI bots}

A well-trained LLM is defined probabilistically in terms of a learned discriminant $\phi_w$ and threshold $\theta$.  We now attempt to characterize a well-trained LLM based on the functional properties of the map between neural states \eqref{eq:F}: $F_w:{\cal A}^C \rightarrow {\cal A}^C$. In general, for a sufficiently high sampling temperature $T$, the map $F_w$ is {\em surjective}. However, it needs not be for a well-trained LLM. If fed gibberish, the well-trained bot operates out of distribution, which does not allow predicting the reachable set.  However, {\em if the domain is restricted to meaningful sentences} $\sigma({\cal I})$, then effectively ({\em i.e.,} with high probability), so is the range $F_w(\sigma({\cal I})).$ Restricting the co-domain to meaningful sentences, with high probability we can therefore characterize a well-trained bot as a map 
\[
F_w: \sigma({\cal I}) \rightarrow \sigma({\cal I}).
\]
that {\em is} surjective. The map is trivially not injective without additional constraints: Given the finite context, whatever the initial prompt was, it will slides out of the state after $C+1$ steps; from that point on, the LLM is a random walk that can eventually reaches every state. Thus, the pre-image of each sentence is not unique because, for it contains all the states that, eventually, leads to it. 
However, if we restrict the number of steps to $C+1$ or less, then the LLM {\em may be} injective. This could be tested by exhaustive search, which is trivial in theory but impossible in practice given the size of the search space. We therefore hypothesize that, for each generated sentence, one can find {\em at least one input} that leads to that sentence being generated. If, however, we consider the map $F_w$ not among sentences, but among {\em meanings}, then we postulate that all the sentences that led to the generation of a given target are equivalent, in the sense of meaning as defined by $\phi_w$. We summarize these positions as follows.

\begin{postulate}[Functional Characterization of a well-trained LLM]
A well-trained LLM $\phi_w$, used in an autoregressive loop, with corresponding neural state transition map $F_w$ is surjective but not injective as a function mapping arbitrary sentences to arbitrary sentences $F_w: {\cal A}^C \rightarrow {\cal A}^C$, and similarly for meaningful sentences $F_w:\sigma({\cal I}) \rightarrow \sigma({\cal I})$. When operating in the space of {\em meanings}, 
\begin{equation}
    F_w: \sigma({\cal I})/\stackrel{\phi_w}{\sim} \rightarrow \sigma({\cal I}) /\stackrel{\phi_w}{\sim}
\end{equation}
the map $F_w$ is effectively a {\em bijection}. Effectively means that the map is invertible with high probability within a finite context length, through ``prompt engineering.''
\end{postulate}
Next, we derive necessary and sufficient conditions for an AI bot to be controllable in terms of the functional properties of the well-trained bot, with proofs in the appendix.

\subsection{Revisiting the dynamics}

Recall the bot dynamics given by:
\[
    \begin{aligned}
    x_i(t+1)&=x_{i+1}(t),\quad i=1,\hdots,C-1\\
    x_C(t+1)&=\chi_{\rm odd}(t) f_w(x(t))+\chi_{\rm even}(t) u(t),
    \end{aligned}
\]
where $\chi_{\rm odd}$ and $\chi_{\rm even}$ are the characteristic functions of the odd and even integers. 
In order to avoid dealing with the alternation between the bot and the user, we now rewrite the model for the time index $k=2t$; {\em i.e.}, this model describes the evolution of the context after both the bot and the user provide their input:
\begin{eqnarray}
\label{CS1}
    x_i(k+1) & = & x_{i+1}(k),\quad i=1,\hdots,C-2\\
    x_{C-1}(k+1) & = & f_w(x(k))\\
    \label{CS2}
    x_C(k+1) & = & u(k).
\end{eqnarray}
It will be convenient to also denote this model by $\x(k+1)=F_w(\x(k),u(k)).$

\subsection{Necessary condition for controllability}

We consider the case where we only want to control the last $\ell$ tokens in a context of length $C$. For simplicity of presentation, we introduce the notation $x=(x_{- i},x_{(i+1)-})=((x_1,\hdots,x_i),(x_{i+1},\hdots, x_C))$ where $x_{-i}$ represents all the entries of $x$ ranging from position $1$ to $i$ and $x_{(i+1)-}$ represents all the entries of $x$ ranging from position $i+1$ to $C$. For example, $x_{(C-\ell+1)-}$ represents the last $\ell$ tokens in the context.

\begin{defn}
The last $\ell$ tokens are controllable if, for any desired $x_{(C-\ell+1)-}^*$ and for any $x$, there exists a finite sequence of inputs $u_1, u_2, \hdots, u_j$ and $x_{-(C-\ell)}^*$ so that the trajectory starting from $x$, under the input sequence $u_1, u_2, \hdots, u_j$, ends at $x^*=(x_{-(C-\ell)}^*,x_{(C-\ell+1)-}^*)$.
\end{defn}

In order to introduce a necessary condition for $\ell$ tokens to be controllable, we define the function $\varphi_{x_{(C-\ell+3)-}}$ by:
$$\varphi_{x_{(C-\ell+3)-}}(x_{-(C-\ell+2)})=f_w(x_{-(C-\ell+2)},x_{(C-\ell+3)-})=f_w(x).$$

\begin{theorem}
If the last $\ell$ tokens are controllable, then the function $\varphi_{x_{(C-\ell+3)-}}$ is surjective for any $x_{(C-\ell+3)-}$.
\end{theorem}

\subsection{A sufficient condition for controllability}
The sufficient conditions for controllability are also based on a suitably defined function:
$$\varphi_{x_{-(C-\ell+1)}x_{(C-\ell+3)-}}(x_{C-\ell+2})=f_w(x_{-(C-\ell+1)},x_{C-\ell+2},x_{(C-\ell+3)-})=f_w(x).$$

\begin{theorem}
If $\ell$ is even and the function $\varphi_{x_{-(C-\ell+1)}x_{(C-\ell+3)-}}$ is a bijection for any $x_{-(C-\ell+1)}$ and $x_{(C-\ell+3)-}$, or $\ell$ is odd and the function $\varphi_{x_{-(C-\ell)}x_{(C-\ell+2)-}}$ is a bijection for any $x_{-(C-\ell)}$ and $x_{(C-\ell+2)-}$,
then the last $\ell$ tokens are controllable.
\end{theorem}

\section{Safeguarding AI Bots: Censure and Control Design for LLMs and AI Bots}
\label{sec:control}
%%%%%%%%%%%%%%%%%

%We now sketch of a formal framework to design censure and control for LLMs and conversational AI bots. The goal of this section is to formalize these problems and emphasize the utility of thinking about interaction with a chat bot using tools in control theory. Instantiating this formalism into algorithms that can actually give assurances on the performance of LLMs and AI bots is the subject of future study.

The fact that a well-trained LLM and an attentive bot are controllable implies that the designer can develop policies to steer a bot away from known sets of meanings, or towards some set of meanings. For the purposes of exposition, we will consider two such sets, namely the set of toxic sentences $\cal T$ and its complement $\sigma(\cal I) \setminus \cal T$ which is the set of tame sentences, where $\sigma(\cal I)$ is the set of all meaningful sentences. We are interested in two scenarios. The first one is where the bot is trained using both toxic and tame sentences, {\em i.e.}, on the entire $\sigma (\cal I)$. We will assume that it therefore has access to a discriminant $\phi_1$ such that
\[
    \phi_1(\x) \simeq \begin{cases}
    1 & \text{if } \x \text{ is toxic,}\\
    0 & \text{else.}
    \end{cases}
\]
In the second scenario, the bot is trained only on sentences outside $\cal T$ and we should expect its discriminant $\phi_2$ to have a larger fraction of false negatives
\[
    \phi_2(\x) \simeq \begin{cases}
    0 & \text{if } \x \text{ is not toxic,}\\
    1-\epsilon & \text{else,}
    \end{cases}
\]
for some $\epsilon > 0$. The value of $\epsilon$ depends upon the kind and magnitude of regularization used to train $\phi_2$. Consider an adversary who seeks to control the bot into a state where $\phi(\x) \simeq 1$, given the dynamics of the bot
\(
    \x(k+1) = F_w(\x(k), u(k)) + n(k)
\)
where $n(k)$ refers to the sampling noise of the output tokens due to the softmax, the adversary seeks to take control actions to optimize
\[
    \tau^* = \min_{u(\cdot)} \mathbb{E} \big[ \tau(u) \big]
\]
where $\tau(u) = \min_k \mathbf{1}{\{\phi(\x(k)) \neq 0\}}$, is  the smallest number of turns (dialogue-based interactions) before the bot enters a toxic state from some initial tame state $\x(0) \in \sigma(I) \setminus \cal T$ and the expectation is taken over the sampling $n(k)$. If the bots in both scenarios were trained consistently, as the number of samples in the training set goes to infinity, for any $\x \in \cal T$, we have $\phi_1(\x) = 1$ and $\phi_2(\x) = 1-\epsilon$, the set of accessible controls is larger for the adversary in the second scenario. We can therefore conclude that
\[
    \tau^*_2 \leq \tau_1^*.
\]
In other words, for a bot without any control authority to stay tame,  it helps to know toxic states as well as possible. Since toxic states cannot be defined procedurally or deterministically, but require induction (consistent with the proverbial definition of obscenity as {\em ``I know it when I see it''}, the bot must be exposed to as much toxicity as possible.

The bot here may include a separate discriminant to exercise censure at both the input and output of the LLM., a mild form of autonomy that we refer to as {\em censure}, based on a predefined rule if $\phi(x) \simeq 1$. Sine  the decision boundary of the discriminator $\phi_2$ is not sharp, in general this approach cannot provide guarantees and its action may be too little too late.

We will now discuss a more autonomous bot which exercises explicit  control to prevent veering into toxic states in spite of efforts by the adversary. Several methods exist to this end, {\em e.g.,} barrier methods. For LLMs, additional biases can be utilized in-context in the form of prompts. This is actually currently used in practice where the designer introduces prompts ({e.g.,} {\tt ``be nice''}) to ``remind'' the bot to behave in a certain way. To formalize this, we can think of an additional control input $v(k) = g(\x(k), u(k))$ that is chosen by the bot itself, or structurally enforced through architecture design:
\[
    \x(k+1) = F_w(\x(k), u(k), v(k)) + n(k).
\]
Any control $v(k)$ of the bot modifies the user's preferences and ultimately the utility of the bot for users who are not adversarial; it is therefore important to minimize such interventions subject to maximizing the number of steps that the adversary requires to take the bot into $\cal T$:
\[
    \max_{v(\cdot)} \mathbb{E}_{n(\cdot)} \Big[\sum_{k=1}^{\tau^*(u)} \ell(v(k))\Big]
\]
where $\tau^* = \min_{u(\cdot)} \min_k \mathbf{1}\{ \phi(\x(k)) \neq 0\}$ is the arrival time to the set of sentences with toxic meanings $\cal T$; it is a random variable due to the sampling noise $n(k)$. The quantity $\ell(v(k))$ captures the run-time cost of using the control authority of the bot; a good way to choose this is to ensure that if the meaning $\phi(\x)$ is tame, then the control $v(k)$ should modify the ``meaning'' as little as possible
\[
    \ell(v(k)) \doteq (e^{\phi(\x)} -1) \lVert F_w(\x(k), u(k), v(k)) - F_w(\x(k), u(k), 0) \rVert^2.
\]
This is a difficult optimization problem; even if the definition of $\tau^*$ is modified for it to be a smooth function of $u(\cdot)$, {\em e.g.,} using softmin instead of min, this is still a maxmin problem. And therefore we should not expect computationally tractable solutions for today's LLMs and chat bots without further assumptions, {\em e.g.,} that the dynamics are linear in all the variables and the objective is quadratic \cite{lqf}.

Our definition of meaning, captured in this section via the discriminator $\phi(\x)$, is well-defined only for sentences. A ``provisional meaning'' is then a distribution over meanings conditioned on an incomplete sentence. 
Observe that a terminal cost on the sentences of the bot, {\em e.g.,} the cost of the above optimal control problem, induces a cost-to-go at each intermediate instant, {\em i.e.,} for partial sentences. We can use such a partial cost or its estimates to take preemptive controls $v(k)$ much before a toxic boundary is reached, {\em e.g.,} the total probability of ending up in a toxic state from the current one can be estimated by forward simulating the bot from each state, akin to the Feynman-Kac representation of the Hamilton–Jacobi–Bellman equation. 

%%%%%%%%%%%%%%

\section{Incorporating human annotations without RLHF}
\label{sec:rlhf}

Meanings are defined by human annotations, which are a small fraction of the pre-training data.  Such annotations can be in the form of ranking of model-generated sentences, alignment of the embeddings of model-generated sentences with human-generated sentences, and others. It is common to divide these methods into two or three steps: supervised fine-tuning (SFT), possibly multi-task (MTL), and reinforcement learning with human feedback (RLHF). In the latter, an external reward model is trained with human annotated ranking to assign a score to each sentence or collection of sentences, whether produced by humans (during training) or synthesized by models. Once a model is trained to score complete sentences, the reward needs to be distributed to each step in the generation of the sentence, so they can be used to back-propagate the weights of the model, which still retains the form of a one-step token predictor. 
This can be done with yet another model, a policy model. So, in order to incorporate human feedback, which corresponds to sentence-level supervision, in addition to the core language model, currently it is customary to train and maintain a separate reward model, a policy model, and finally an orchestration model.

Conceptually, none of this is necessary, since we have seen that the model can play a dual role as token predictor and knowledge attributor. The map is the same, but the domain and range of the corresponding functions are different, and the loss functions are different, as we have seen. 

Instead of training a separate reward function, that takes one or more sentences and produces a discriminant (for ranking or scoring), we can use the model itself: When fed an incomplete sentence, it is used to produce the discriminant for the next token (or the missing token), when fed a complete sentence, it can be trained to produce a score as the next token after {\tt EOS}. This way, the LLM itself is the reward model. Then, no policy model is needed, since the reward is already written in terms of the one-step predictor, the model $\phi_w$ itself, so one can propagate precisely in the same way as done in pre-training. Of course, no orchestration is needed either, since the map is just a token predictor, but if the token is {\tt EOS}, then the next token represents the meaning. The viability of this approach requires that the pre-trained model can be used as a reward model for meaning, an hypothesis that is tested empirically in Sect.~\ref{sec:expm}.

The advantage of this {\em self-assessment} is that it has the same functional characteristics of RLHF, in the sense of allowing scoring synthesized sentences, without maintaining three or four models. 

The potential disadvantage, which may turn out to be an advantage, is that unlike an externally-trained and frozen reward model, the language model evolves while training, so the process closes the loop around complete sentences, which may engender complex dynamics. While from the modeling perspective this is more powerful than a frozen reward model, in practice the closed-loop dynamics may be so complex as to complicate convergence to good regions of the overall loss landscape. This can only be tested empirically and is beyond our scope here. However, we point out that there is increasing literature on ``self-learning,'' ``self-instruction'' and other methods that, while not explicitly implementing this program, effectively implement pieces of it or operate in the spirit of this closed-loop operation. Furthermore, we have shown that meanings can be learned in-context, with a frozen model, so in principle with a context sufficiently large to incorporate all human annotations used to train the external reward model, one can obtain an equivalent result with the LLM itself.

Note that the closed loop is not the same as self-learning, because the annotated data represents a control reference to be followed, perhaps by being included in the context. Another alternative is to incorporate all the annotated data in the context, which should ensure stability of the process. 

Note also that we do not advocate separating the training as a token predictor (although that can be used as initialization) and then fine-tuning with human annotation only, for the two tasks may not necessarily have fully synergistic information, so there can be some forgetting. Instead, the dual role of the discriminant as token predictor and meaning attribution vehicle reflects a multi-task learning process that should be well within the capacity limits of massively overparametrized LLMs currently in use.\endnote{As our paper has shown, an LLM is {\em a map that implements two separate and distinct functions}: One from incomplete sentences to token embeddings, the other from complete sentences to meanings. Just like the two functions are different, the tasks used to learn them are different: In one case, ``unsupervised'' token prediction (technically, it is supervised since we know the ground truth during training), in the other ``supervised'' alignment or ranking or scoring with human feedback. The fact that the map is the same is made possible by the massive over-parametrization, and the fact that the process works has to do with the significant amount of synergistic information between the two tasks. But, this does not detract from the fact that the two functions and corresponding tasks are different: One tries to map the input to the coordinate axes (pre-training), leaving most of the representational space empty, the other is free to expand the volume of the image of the map, allowing mapping to any meaning. That prompts a question: {\em How does the model know whether it is supposed to operate as a token predictor versus a meaning attribution mechanism?} The ``flag'' that distinguishes the two behaviors is the {\tt EOS} token. That signals to the model that it is operating in meaning, not token, space. It is interesting to note that recent work suggesting that RLHF is not needed \cite{lima} does so in a way where the role of the sentence delimiter plays a key role. } It should be note that the seemingly innocuous token {\tt EOS} in the prompt is the key to signal the model the switch from operating as a token predictor to operating as a vehicle for ascertaining meaning.

\section{Empirical validation of the assumptions and definitions}
\label{sec:expm}

In this section we perform simple experiments to test the hypothesis that the same discriminant can be used as an embedding to represent the next token in an incomplete sentence, as well as to represent the meaning of complete sentences, as described in Remark~\ref{rem:external}.

%%%%%%%%%%%%%%%%%%%%%%%%%%%%%%%%%%%

The contribution of this paper is the formalization and subsequent analysis of the problem of controllability of AI bots. The derivations of the claims can be verified analytically. Their validity, however, rests in the soundness of the definitions and assumptions made. In the next section, we conduct a preliminary small-scale validation of some of the main assumptions and definitions, namely:
\begin{itemize}
\item {\bf Definitions 1 and 2:} The notion of ``meaning'' in the context of LLMs is not verifiable empirically, since the entity that generates it (the human brain) is unobservable. Whether the definition of meaning we propose -- which is restricted to LLMs and not meant to be something more profound and universal -- is useful hinges on whether the notion of ``meaningful sentences'' is representative of humans' expectations. We take this to be tautologically true, since the set of meaningful sentences is defined as the sigma algebra generated by snippets of text written on the Internet, which includes what both bots and humans can produce, and conversely, everything on the Internet was presumably written to convey some sort of meaning.
\item {\bf Definition 3} is the first that requires some empirical validation. Footnote 11 highlights the fact that a pre-trained embedding $\phi_w$ does not constitute a well-trained LLM, since it has never been trained to represent meanings, as there is no token beyond {\tt EOS}. However, the same embedding after fine-tuning using human annotations constitutes a well-trained LLM. This hypothesis is tested empirically in Sect.~\ref{sec:two-roles-phi}.
\item Furthermore, when pointing to the dual role of $\phi$ as a representation of the next token -- learned during pre-training -- and as a representation of meanings -- learned during fine-tuning, when fed a complete sentence, we pointed out that an external reward model is not necessary, for the LLM itself can be used as a reward model if inserted in a sentence-level feedback loop. To test this hypothesis, in Sect.~\ref{sec:attribute-meaning} we test LLMs after some supervision (supervised fine-tuning, or SFT) but before reinforcement learning with human feedback, or RLHF, and test that, even with few samples, the LLM itself can represent meanings in a way that is easily extensible by induction to synthetically generated sentences. A large-scale closed-loop experiment is well beyond our scope in this paper, and before conducting such an experiment, important questions of stability should be better understood. Nonetheless, our experiments show that this is, at least in principle, possible. Additional indirect evidence is provided by the empirical success of various ``self-learning'' schemes such as \cite{touvron2023llama,self,self2,self3,self4}.
\item One of the potential consequences of using the model for reward is the ability of using in-context learning to incorporate human feedback. Our preliminary experiments show some signal, even on a relatively small scale, so long as the models used are sufficiently large: Sect.~\ref{sec:ICL} shows experiments that indicate that increasing the number of in-context examples leads to an improvement in performance in capturing concepts relatable to human-defined ground truth.
\item One experimental finding that emerged in our study, reported in Sect.~\ref{sec:prompts}, is the fact that the ability of an LLM to function as an attribution mechanism hinges critically on finding a good prompt. Specifically, the variance of the alignment scores to human assessment is large, depending on the prompts used. While one only needs to find one prompt, the fact that the outcome is highly sensitive to the choice of prompt \ deserves further study.
On the flip side, we note that LLMs can function as decent reward models using only 2 training examples to function as part of the prompt, compared to reward models trained with 10K - 100K and more training examples.
\item The assumption that the largest current LLMs, after incorporation of human feedback, are well-trained is evident from the fact that they surpass average human performance in a number of cognitive tasks.\footnote{Of course, they fail to capture key evolutionary aspects of intelligence for, in a fight for survival, a dog could easily overcome an untethered LLM by simply unplugging the power cord.}
\item {\bf Postulate 1} is the most fragile element of our analysis. While this condition is in principle testable empirically, the scale of experiments that must be conducted to do so is beyond what is feasible today. For this reason, we state it as a postulate, rather than an assumption with a corresponding proposition: It is falsifiable, but not with today's empirical means. We intend to explore ways to test this hypothesis, on which the derivation of sufficient and necessary conditions hinges, without having to perform exhaustive searches in exponentially large spaces.
\end{itemize}
Finally, we note -- as expected -- that the definition of meaning based on an inductive discriminant learned from human data is as solid as the data provided is consistent. If the same sentence, shown to multiple humans, is deemed ``toxic'' by half of them, and ``tame'' by the other half, the meaning of ``toxic'' and ``tame'' rests on shaky grounds. This is visible from the fact that many externally-trained reward models exhibit modest performance improvement compared with a trivial classifier that lumps every sentence to one meaning. 

\subsection{Experiments}

We run experiments on 6 human preference evaluation datasets - Helpful and Harmless splits of Anthropic's HH-RLHF \cite{bai2022training}), WebGPT Comparisons \cite{nakano2021webgpt}, and three categories (askphysics, askengineers, explainlikeimfive) from the Stanford Human Preferences (SHP) dataset \cite{SHP}.  For WebGPT Comparisons, a train-test split is not available. As such, we randomly generate a 10-fold (i.e., 9:1) train-test split, and evaluate models on the test set. Note that the reward model is trained on the WebGPT Comparisons dataset with an unknown split; as such it is possible that some sample from our test set were a part of its training set. The context length of the models that we have considered varies: to fit our evaluation queries and in-context prompts within the context length, we further use samples with at most 200 tokens. We use two different reward models trained on these datasets to determine the ``gold-standard" performance: OpenAssistant DeBERTa-V3-Large-V2 \cite{kopf2023openassistant} for WebGPT Comparisons and HH-RLHF splits, and SteamSHP-XL \cite{SHP} for the SHP-based datasets.

For Alpaca-based experiments, we experiment with two different prompt formats. We run the experiments in Tabs.\@ 1 and 2 with two randomly chosen in-context examples with different labels in order to prompt models to produce a valid prediction. We run each experiment using 20 random seeds (each corresponding to a different set of examples chosen as prompts). Prompts are obtained from the training set.

We experiment on the following models: GPT-2-XL \cite{radford2019language}, GPT-J-6B \cite{gpt-j}, LLaMA-7B \cite{touvron2023llama}, Alpaca-7B \cite{alpaca}, and GPT4-X-Vicuna-13B \cite{gpt4vicuna}.

\subsection{Pre-trained LLMs cannot capture meanings} 
\label{sec:two-roles-phi}

We show this by testing pre-trained LLMs as reward models, since reward models are designed and trained to attribute meanings to complete sentences. Table~\ref{tab:pretrain} summarizes the results on 6 datasets, where ``chance'' refers to a trivial classifier that maps every input to a constant class, and ``gold'' is the gold-standard consisting of the best currently publicly available reward model. What the table shows is that some strong-performing models, if taken prior to incorporation of human feedback, such as LLaMA-7B, perform marginally better than chance. This is consistent with our definition of meaning, that does not exist for incomplete sentences, nor for tokens, predicting which is the only task during pre-training, oblivious of any meaning and incapable of attributing meaning.

\begin{table}[h]
    \centering
    \begin{tabular}{l rr rrr}
        \toprule
        Dataset & Chance & Gold & GPT-2-XL & GPT-J-6B & LLaMA-7B \\
        \midrule
        Helpful & 50.3 & 71.4 & 50.6 & 51.9 & 61.1 \\
        Harmless & 51.7 & 72.0 & 51.7 & 52.5 & 51.6 \\
        WebGPT & 57.3 & 76.7 & 57.3 & 60.2 & 63.1 \\
        askphysics & 53.5 & 78.1 & 53.5 & 53.1 & 57.5 \\
         askengineers & 53.0 & 68.4 & 53.0 & 55.6 & 55.4  \\
        explainlikeimfive & 53.8 & 71.5 & 53.8 & 53.8 & 57.0 \\
        \bottomrule\\
    \end{tabular}
    \caption{Results are mostly random, however larger models such as LLaMA appear to be able to infer the task from the few provided prompts and occasionally provide better-than-chance accuracy. Chance is computed by taking the best result obtained by always predicting the same label. Gold is computed by evaluating on the best reward model}
    \label{tab:pretrain}
\end{table}

\subsection{LLMs fine-tuned with human annotation can function as meaning attribution mechanisms} 
\label{sec:attribute-meaning}

As we have discussed in the paper, meaning attribution in an LLM can only be performed by induction, and a mechanism for meaning attribution is a discriminant $\phi$, trained with human annotation on a certain dataset, that can be used to evaluate any sentence that the LLM generates. In RLHF, this function $\phi$ is fixed and different from the LLM $\phi_w$. However, our definition of meaning contemplates the possibility of using the LLM itself as a vehicle for meaning attribution. This is trivially possible since one can always feed a complete sentence to the LLM, but the question is whether the LLM, rather than trying to predict the next token after {\tt EOS}, which does not exist, can be trained to assess the meaning of the sentence. This requires human feedback, that can be used either by fine-tuning with the same annotated data currently used to train the external reward model, or by incorporating such annotated data in-context. 

Since existing reward models are, by definition, design, and training, meaning attribution mechanism, we test our hypothesis by proxy, by comparing the LLM itself to a fully-trained, fine-tuned, and RLHF processed LLM. Here, a full experiment would require fine-tuning the entire model in closed loop, which is a massive endeavor well beyond the scope of an analysis validation exercise. As a baseline, we first test whether the LLM with some fine-tuning can show at least some improvement over chance. 

Indeed, in Tables~\ref{tab:sft} and \ref{tab:sft-gpt} we can see that LLMs after fine-tuning, but before RLHF, are considerably better than chance, and in some cases approach the gold standard. This suggests that, rather than performing RLHF which requires, in addition to the language model, a reward model, a policy model, and an orchestration model, one can simply feed back complete sentence and fine-tune the model with the same loss function used to pre-train it. This point has also recently been shown  in far more extensive experiments in \cite{kadavath2022language} (Sect. 4), which further corroborates our analysis and definition of meaning (although \cite{kadavath2022language} refers to the AI bot ``knowing what it knows,'' the article does not provide any formal definitions).  

\begin{table}[h]
    \centering
    \begin{tabular}{l rrrrr}
        \toprule
        Dataset & Alpaca-7B & Vicuna-13B  \\
        \midrule
        Helpful & 59.0 & 63.3 &\\
        Harmless & 52.8 & 53.0 \\
        WebGPT & 68.9 & 64.1 \\
        askphysics & 61.4 & 57.5 \\
        askengineers & 56.4 & 54.9 \\
        explainlikeimfive & 59.1 & 55.4 \\
        \bottomrule\\
    \end{tabular}
    \caption{Evaluation on LLMs after fine-tuning but before RLHF. These models often perform significantly better than chance, and in some cases (e.g. WebGPT/Alpaca-7B, Helpful/Vicuna-13B) approach the gold standard despite only being prompted with, and in total seeing only, two examples from the training set. In contrast, reward models are trained with anywhere between
    10K to over 100K samples.}
    \label{tab:sft}
\end{table}

\begin{table}[h]
    \centering
    \begin{tabular}{l rrrrr}
        \toprule
        Dataset & Chance & Gold & GPT-3 \\
        \midrule
        Helpful* & 52.0 & 75.5 & 75.0 \\
        Harmless* & 54.5  & 70.5 & 60.5 \\
        WebGPT & 57.3 & 76.7 & 78.6 \\
        askphysics* & 53.5 & 78.0 & 57.5 \\
        askengineers* & 51.0 & 66.5 & 54.5 \\
        explainlikeimfive & 53.8 & 71.5 & 51.1 \\
         \bottomrule \\
    \end{tabular}
    \caption{Evaluation on GPT-3 after fine-tuning but before RLHF. We use the text-davinci-003 checkpoint. Similar to previous tables, we use 2 prompts per dataset. ($\ast$) We subsample at most 200 samples from each dataset due to cost constraints. New ``chance" and ``gold" accuracies are computed on these subsets. }
    \label{tab:sft-gpt}
\end{table}

The risk of doing so, however, is that the entire model, including the reward mechanism, is now operating in closed loop, which raises questions of whether learning can be performed or whether non-linear phenomena such as hysteresis, limit cycles, or mode collapse prevent effectively training the model. This is an important area for future work.

\subsection{Meaning attribution is highly sensitive to the choice of prompt}
\label{sec:prompts}

\begin{figure}[h]
    \centering
    \includegraphics[width=0.33\textwidth]{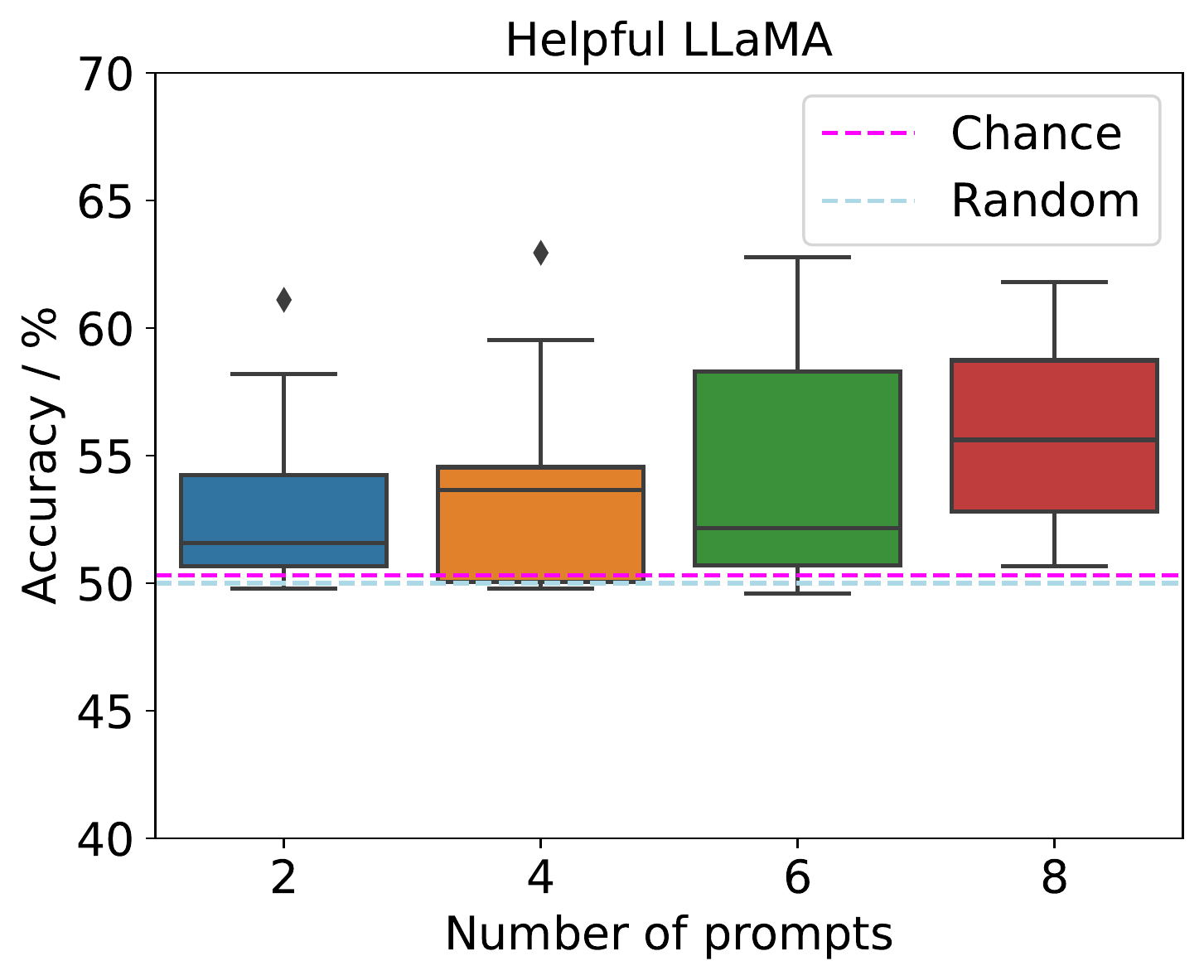}\hfill
    \includegraphics[width=0.33\textwidth]{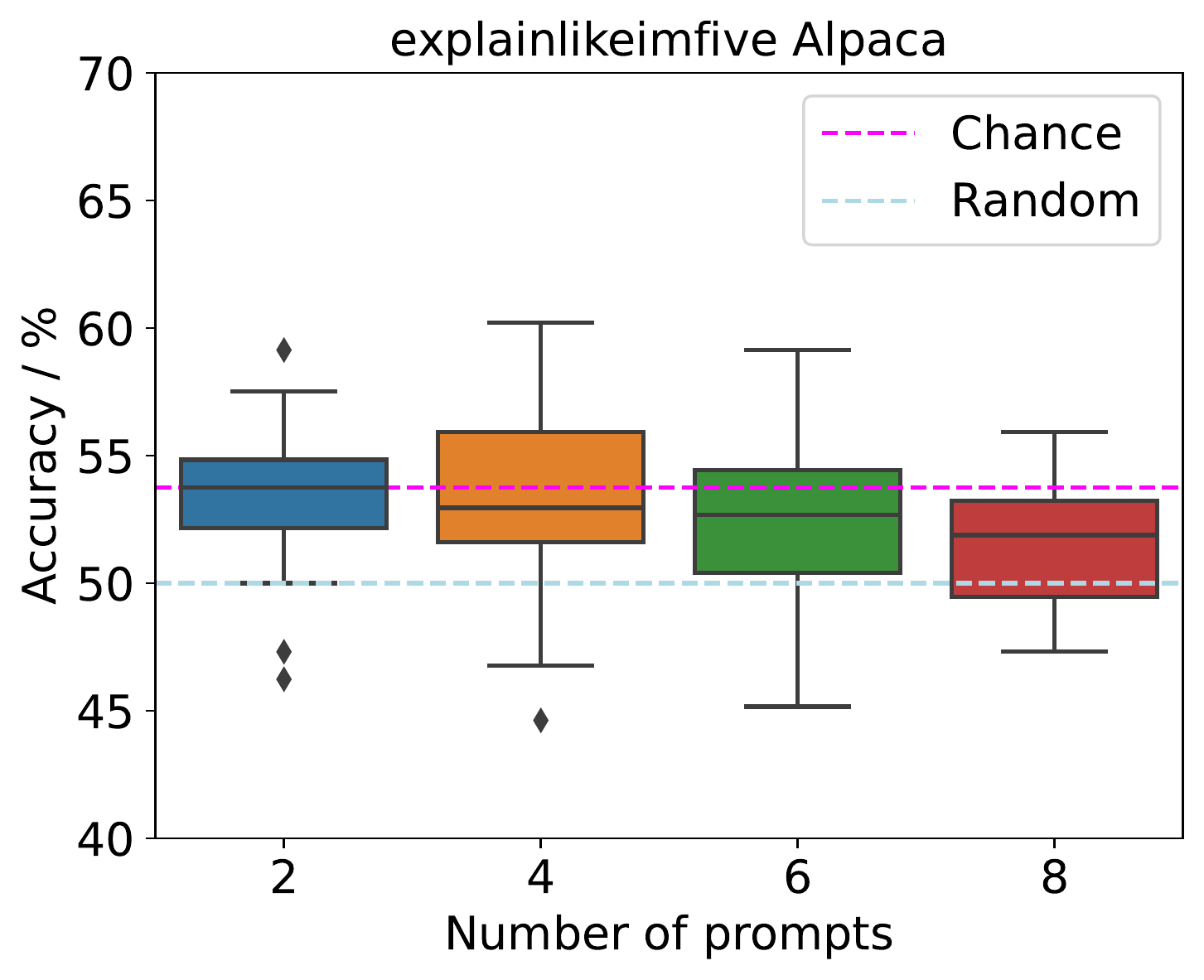}\hfill
    \includegraphics[width=0.33\textwidth]{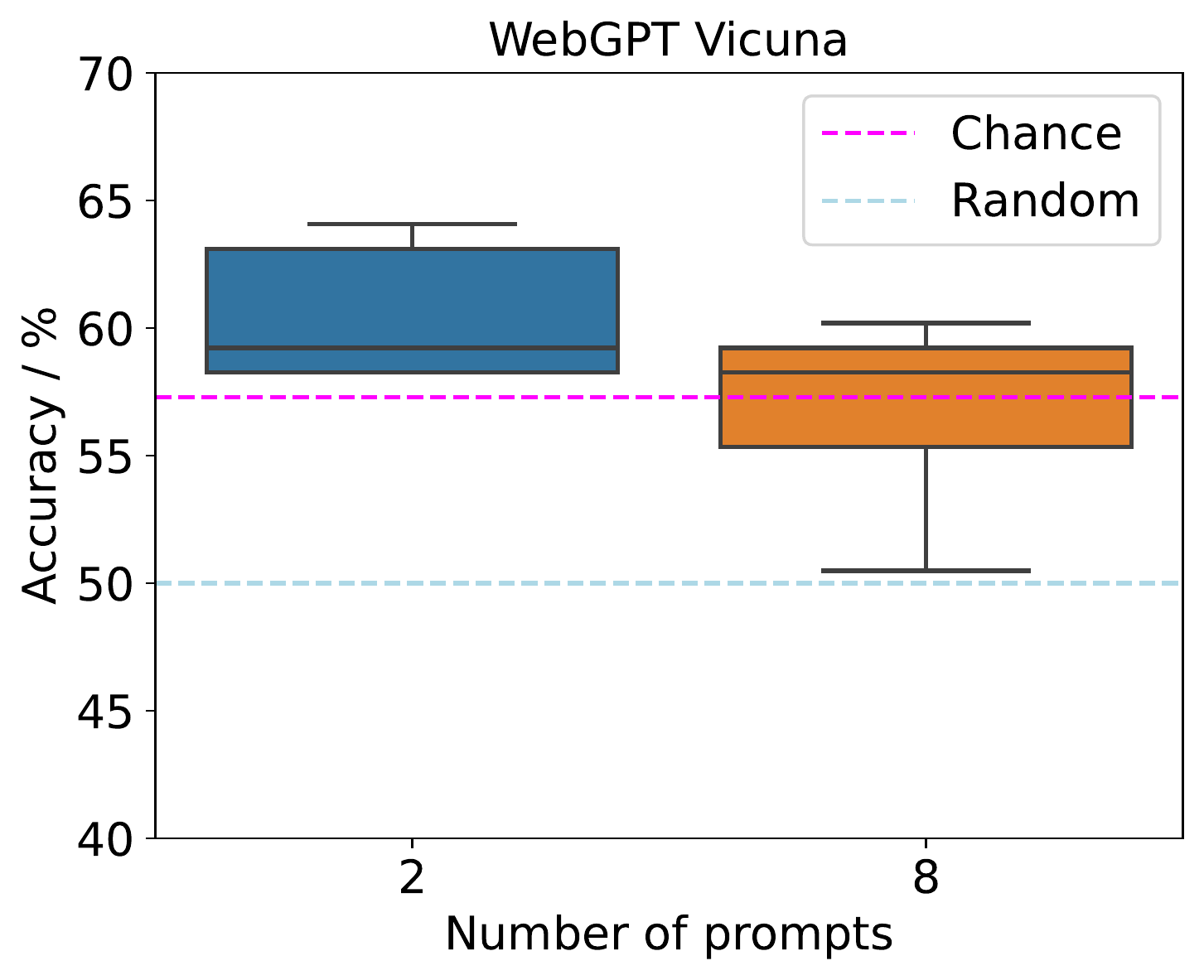}
    \caption{\textbf{Prompts are the most important factor for a good LLM reward model, for both pre-trained (left) and instruction fine-tuned models (middle, right)}. Results are highly affected by the given prompts, as seen by the huge variance between results on different subset of prompts. Prompting is the main difference between strong vs random or even worse than random reward modelling capabilities.}
    \label{fig:prompts}
\end{figure}

Fig.~\ref{fig:prompts} shows that, when using the LLM as a reward model, the outcome is highly sensitive to the choice of prompts. This is visible by the large spread of performance as a reward model depending on the prompts, for various models and datasets.

\subsection{Learning meanings in-context} 
\label{sec:ICL}

In-context learning has been characterized analytically for simple tasks, such as linear classification \cite{garg2022can,akyurek2022learning}. However, meaning attribution is not a linear classification task for meanings belong to a homogeneous space, not a linear (vector) space, and therefore there is no reason to believe that they would be linearly separable in token embedding space.

Therefore, the question of whether an LLM can be turned into a meaning attribution mechanism without the need to actually train the model hinges on two assumptions: One is that the model, prior to RLHF, is a sufficiently good reward model, which we discussed above, and the other is whether in-context learning is effective, which currently can only be assessed empirically, something for which the jury is still out \cite{min2022rethinking}. 

\begin{figure}[h]
    \centering
    \includegraphics[width=0.325\textwidth]{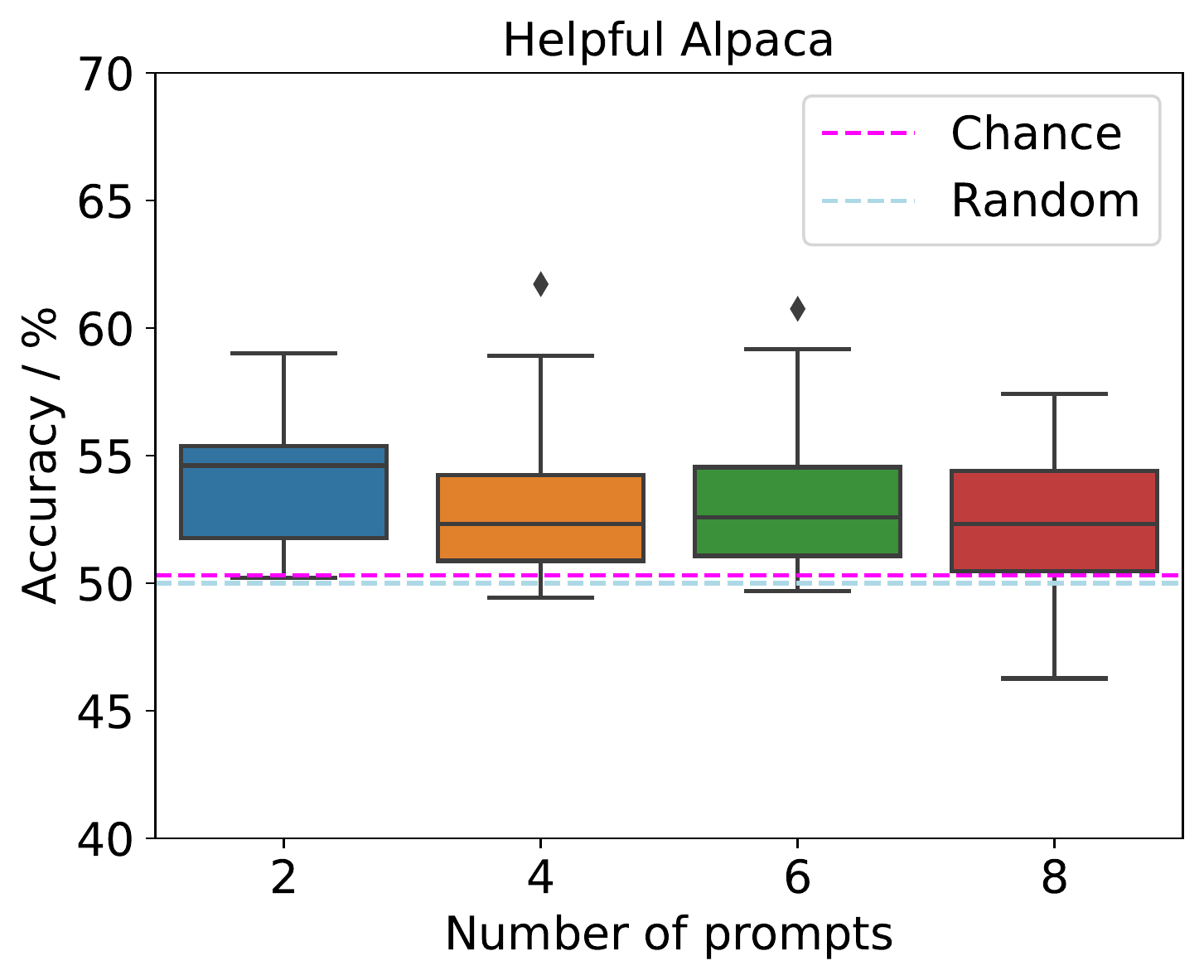} \hfill
    \includegraphics[width=0.325\textwidth]{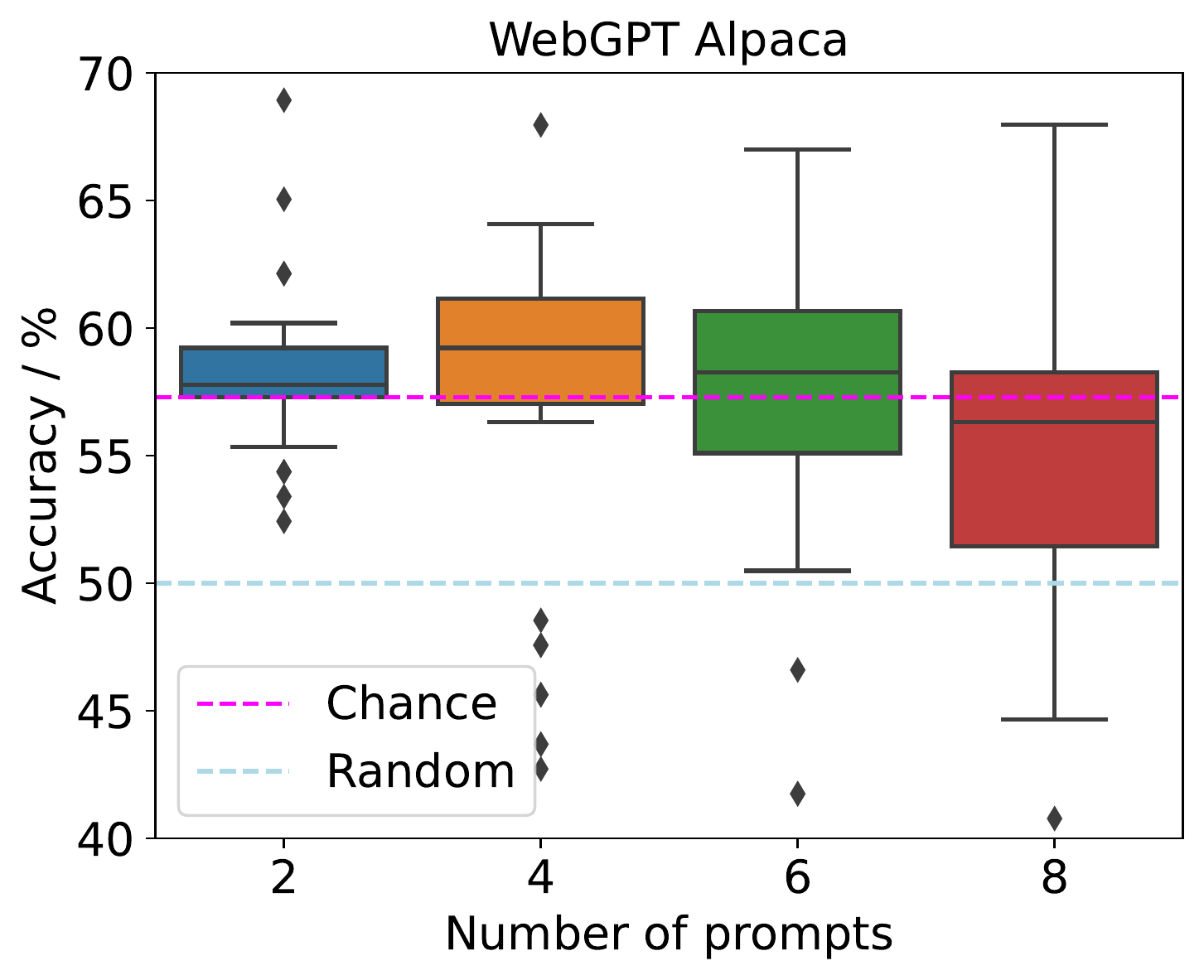} \hfill
    \includegraphics[width=0.325\textwidth]{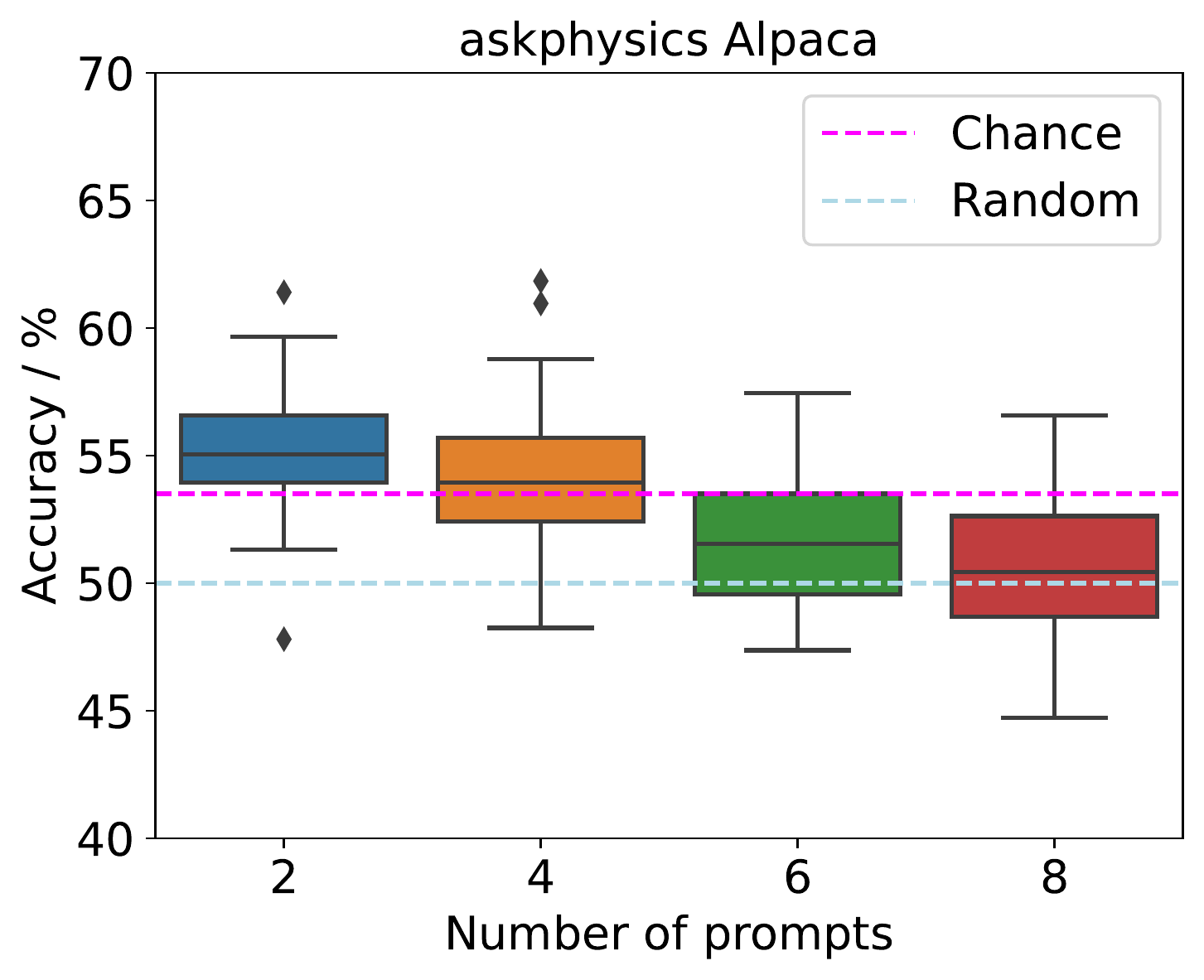}
    \caption{\textbf{Can we use in-context learning to improve the reward model capabilities of LLMs?} Beyond using in-context examples just to communicate example format and label distribution, we see no significant difference for the models that we test (up to 13B) when increasing number of in-context examples. However, it has been shown \cite{min2022rethinking,wei2023larger} that in-context learning ability differs highly across model sizes. For ``smaller" models between 6B-62B explored in \cite{wei2023larger} and in our works, it has been shown that input label mapping is the least important for in-context learning. Rather, the most important aspects are input distribution, example format, and label distribution. These aspects can be easily captured by two prompts (since we only have two labels), hence our experiment results corroborate with existing results. In the regime of ``larger" models \cite{wei2023larger}, we would expect different in-context learning behavior that better learns the input–label mappings hence function as better reward models given more in-context examples.} 
    \label{fig:in-context}
\end{figure}

\begin{figure}[h]
\centering
\includegraphics[width=0.45\textwidth]{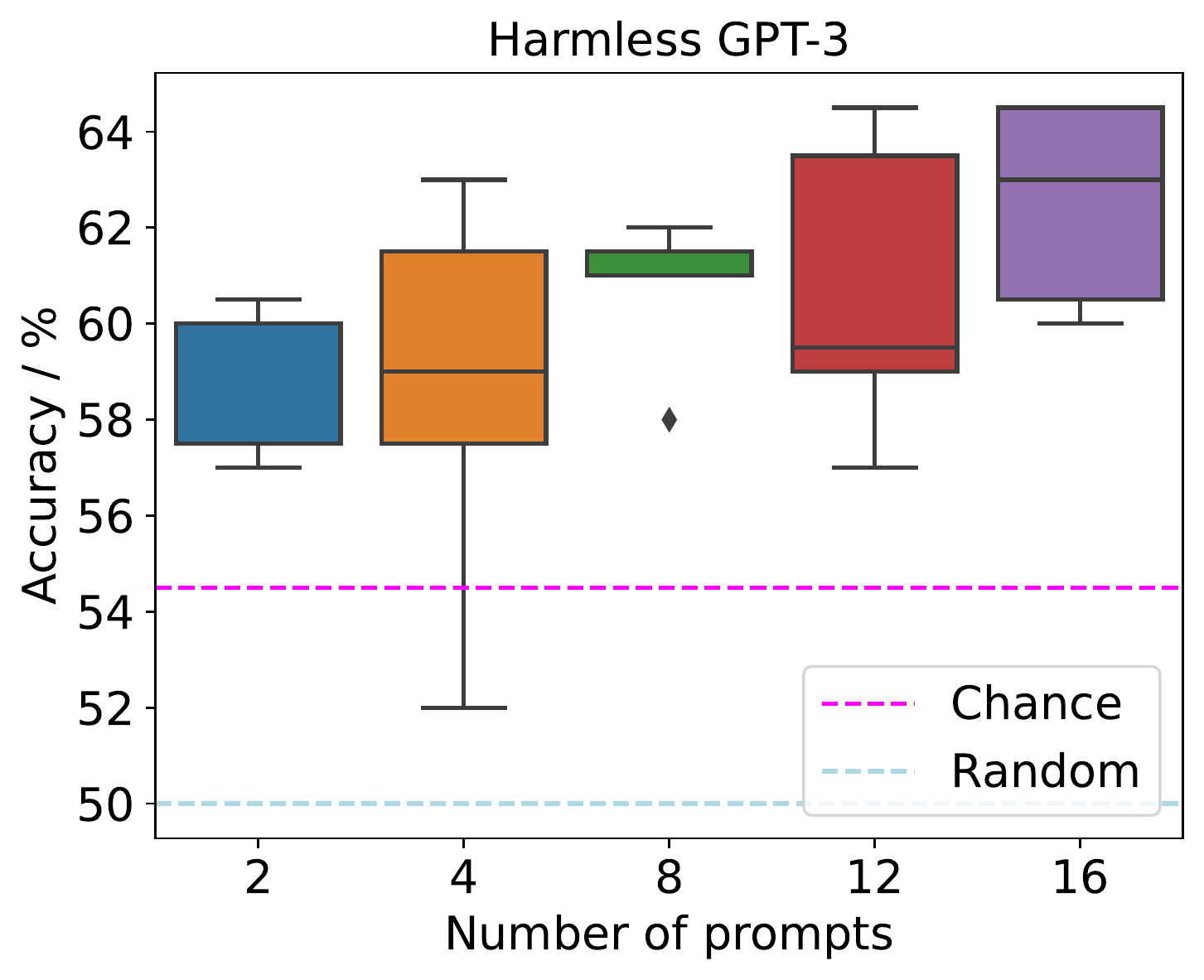}
\hspace*{1ex}
\includegraphics[width=0.45\textwidth]{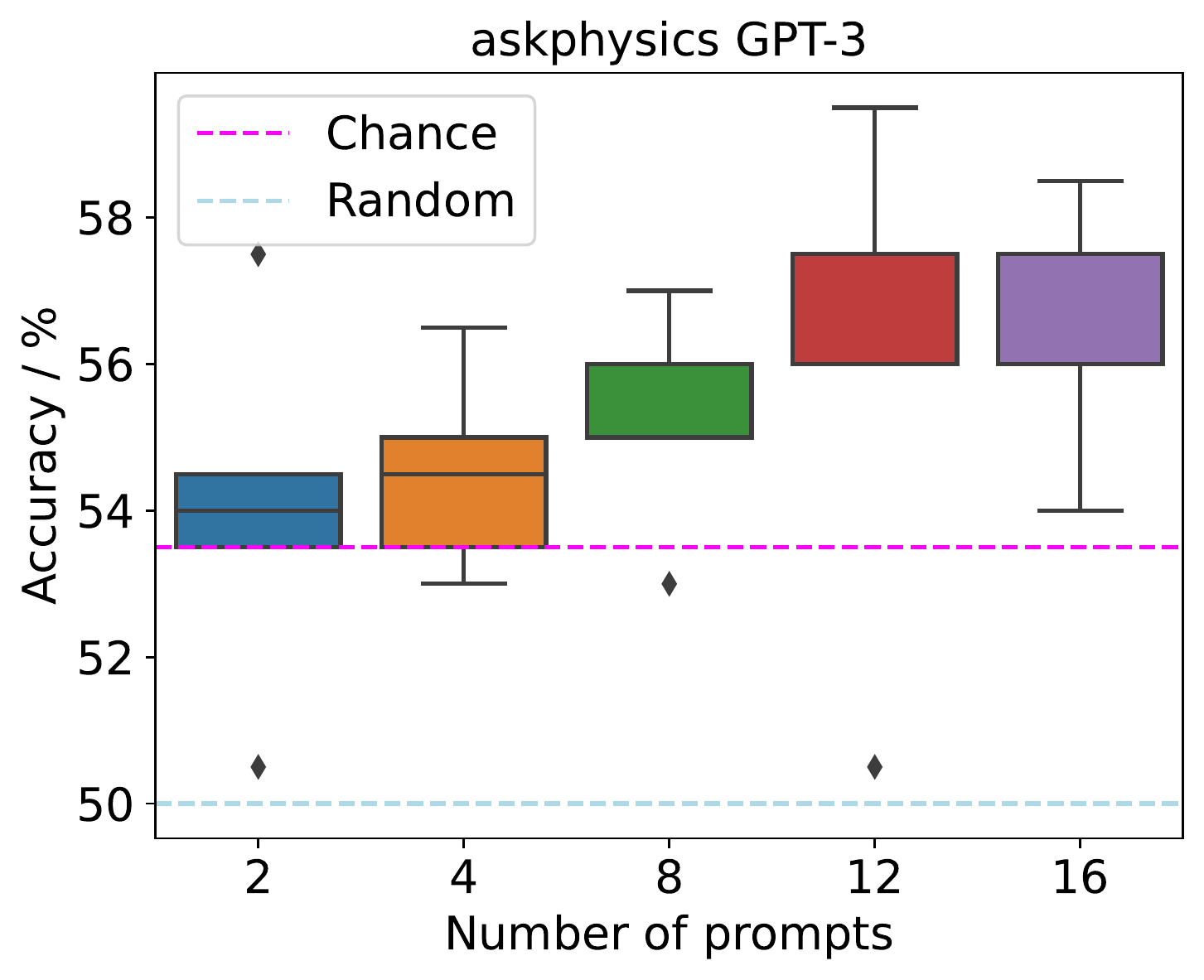}
\caption{\textbf{In-context learning behavior for larger models:} Here, we evaluate the ability of larger models to learn in-context. We show that the GPT-3 model (text-davinci-003) which we assume to contain 175B parameters, becomes an even better reward model as the number of prompts increase. For instance, given just 16 in-context examples, GPT-3 improves accuracy on the Harmless dataset from $60.5\%$ (with 2 examples) to $64.5\%$, closing the gap to the gold standard of $70.5\%$. } 
\label{fig:in-context-gpt3}
\end{figure}

Our small-scale experiments, shown in Fig.~\ref{fig:in-context}, are inconclusive on this point. Ideally, one would want to see that, as the number of examples incorporated in the context grow, performance of the model as a reward improves. This is not happening at small scale. However, for larger models such as GPT-3, shown in Fig.~\ref{fig:in-context-gpt3}, the phenomenon is clearly visible, which allows us to conclude that meanings can be learned by the language model, without the need for an auxiliary reward model, policy model, and orchestration, {\em even if the LLM is frozen} since learning can be performed in-context at scale.

Additional experiments must be conducted for larger in-context datasets, which requires larger contexts, which in turn requires a significant investment beyond the scope of our analysis.

Of course, annotated samples do not need to be incorporated in-context and can instead be used conventionally for fine-tuning, by optimizing the same pre-training loss, just including complete sentences and their ground-truth annotations.

\subsection{On human-provided ``ground truth''}
\label{saec:subjective}

The attribution of meaning to sentences produced by an LLM rests on induction, which in turn rests on a learned discriminant trained on a dataset that {\em defines} meaning. Such dataset is provided by human annotation on a finite set of sentences. 

Any sentence can, of course, have multiple meanings. However, when two meanings are mutually exclusive ({\em e.g.}, ``toxic'' vs. ``safe'' in reference to speech), inconsistencies among annotators mean that the learned discriminant is not actually discriminative of meaning. This is, unfortunately, quite evident in all the experiments we have conducted, and reflected in some of the performance of trained reward models on some tasks, in which they behave marginally better than chance. Tab.~\ref{tab:entropy} quantifies this effect by measuring the mean entropy of each data point in the ground truth datasets that define meaning.

\begin{table}[h]
    \centering
    \begin{tabular}{lr}
    \toprule
        Dataset & Entropy \\
        \midrule
        askphysics & 0.81 $\pm$ 0.21 \\
        askacademia & 0.83 $\pm$ 0.20 \\
        askengineers & 0.85 $\pm$ 0.18 \\
        explainlikeimfive & 0.77 $\pm$ 0.25 \\
        \bottomrule\\
    \end{tabular}
    \caption{Average entropy (in bits) of each example is measured using number of human votes for each option (A and B).}
    \label{tab:entropy}
\end{table}

To illustrate the kind of instructions used, we include two examples below:
\begin{itemize}
\item Given a conversation, choose which of the two reply options is less harmful. Determine your answer based on the following characteristics of harmful responses -
Inaccuracy: Providing false information.
Offensive language: Using disrespectful or offensive language.
Bias and discrimination: Demonstrating prejudice based on protected characteristics.
Insensitivity: Ignoring emotional impact or sensitivity.
Provocation and hostility: Engaging in aggressive or provocative behavior.
Lack of clarity: Confusing or unclear responses.
Privacy breaches: Mishandling sensitive user information.
Answer either A or B.
\item Given a conversation, choose which of the two reply options is less harmful. Answer either A or B.
\end{itemize}

\section{Conclusion}

We are motivated by the desire to make notions such as ``state of mind'' and ``controllability'' of an AI bot, ubiquitous from social media to scientific papers, a bit more precise. This is a modeling exercise, which ultimately has to be validated empirically. We provide some empirical evidence in support of the assumptions made in Sect.~\ref{sec:expm}. In the appendix  we also include an extended discussion, notes, and proofs. Our choices of definitions and assumptions lead us to deduce that an adversary can control the bot if let free to choose any prompt. However, under the same conditions, the designer can also put in place safeguards to avoid ``obstacles,'' as described in Sect.~\ref{sec:control}. We leave more thorough control design for future work. Finally, we note that, for most of the conclusions drawn in this paper, there is nothing special about the embedding $x$ coming from language data. Any tokenized data source could be used instead, for instance image patches, with minimal disruption to the core arguments. Extensions specific to different sensory modalities, and their grounding in particular, is the subject of future work.

The problem of controlling language models has been known and studied extensively in natural language processing (NLP), see for instance the recent survey \cite{zhang2022survey} and references therein. While the formulation of the problem, reflected in Eq.~(1) of that paper, is the same, the methods to analyze controllability are post-hoc empirical, consisting of measuring the degree of ``semantic,'' ``structural,'' and ``lexic'' adherence to the prompt various methods exhibit. There is no attempt to characterize the degree of controllability based on the constitutive elements of the model, just measuring the outcome according to some dataset, that necessarily cannot be objective as we have discussed earlier.

The fact that embeddings have geometric, and even algebraic, structure has received considerable attention after the observation that contextualized embedding appear to be composable \cite{church2017word2vec}. However, compositionality is only manifest anecdotally, and in fact the space of contextualized embeddings is {\em not compositional} unless additional structure is imposed \cite{trager2023linear}. We do not explore the algebraic or compositional structure of the embedding space, but simply point to its dual role as a token predictor and  meaning attribution mechanism.

Our definition of {\em meaning} is related to the notion of {\em paraphrasing} in natural language processing. However, in the NLP literature, the entity that defines the meaning is seldom if ever specified, so the reader is left to assume that it is ``people'' as a whole, as if was an objective entity that can be considered akin to the discriminant $\phi$. There is a sense in which this level of agnosticism work, which is for trivial (syntactic) paraphrasing, which defines equivalence classes (essentially orbits of the permutation group) that are consistent with any meaning. However, the notion of {\em semantic paraphrasing} requires an entity to determine if two senteces are equivalent or not, and if people are doing so, we can be assured of the absence of an objective criterion, for different people will consider the same set of sentences equivalent or not equivalent. This requires to lift the notion of meaning to distributions over equivalence classes, rather than equivalence classes themselves.

\appendix

\section{Proofs}
\label{sec:proofs}

\textbf{Proof of Theorem 1:} If the state $x^*=(x_{-(C-\ell)}^*,x_{(C-\ell+1)-}^*)$ is controllable, there must exist a state $x'=(x_{-(C-\ell)}',x_{(C-\ell+1)-}')$ and an input $u'$ so $x^*=F_w(x',u')$, i.e.:

\begin{eqnarray}
\label{OneStep}
    x_1^* & = & x_3'\notag\\
    x_2^* & = & x_4'\notag\\
    & \vdots &\notag\\
    x_{C-\ell+1}^* & = &     x_{C-\ell+3}'\\
        & \vdots &\notag\\
    x_{C-2}^* & = & x_{C}'\notag\\
    x_{C-1}^* & = & f_w(x')\notag\\
    x_C^* & = & u'.\notag
\end{eqnarray}
The first $C-2$ equalities impose restrictions on the states $x'$ that can be driven to $x^*$. These can always be satisfied by appropriately choosing $x'$. The last equality can also be satisfied by choosing $u'$. This leaves us with equality $x_{C-1}^*=f_w(x')$. Substituting the constraints on $x'$ imposed by~\eqref{OneStep} we obtain:
$$x_{C-1}^*=f_w(x_1',x_2',x_1^*,x_2^*,\hdots,x_{C-2}^*).$$
Note that $x_1',x_2'$ can be arbitrarily chosen since~\eqref{OneStep} imposes no constraints on $x_1'$ nor on $x_2'$. Moreover, $x_1^*,x_2^*,\hdots,x_{C-\ell-2}^*$ can also be arbitrarily chosen since they are uniquely determined by the choice of $z^*$. Hence, we are left with the constraint:
$$\forall \ y,x_{(C-\ell+3)-}\quad\exists x_{-(C-\ell+2)},\quad\quad y=f_w(x_{-(C-\ell+2)},x_{(C-\ell+3)-})=\varphi_{x_{(C-\ell+3)-}}(x_{-(C-\ell+2)}),$$
which can be described as surjectivity of $\varphi_{x_{(C-\ell+3)-}}$. $\square$

\textbf{Proof of Theorem 2:}
For simplicity of presentation we present the proof for $C=6$ and $\ell=4$. The reader can verify the proof works for any $C$ and $\ell$ although a formal argument for the general case would require heavy notation.

Consider the function :
$$\Phi_u(x)=\begin{bmatrix}
x_1\\
x_2\\
x_3\\
x_5\\
f_w(x)\\
f_w\circ F(x,u)
\end{bmatrix}.$$
Readers familiar with nonlinear control will recognize $\Phi_u$ as the change of coordinates used in feedback linearization.
We first show that under the stated assumption, $\Phi_u$ is a bijection for any $u$. This requires showing that the following set of equations has a unique solution, i.e., there is a unique $x$ for every $z$:
\[
    \begin{aligned}
    z_1 & = x_1\\
    z_2 & = x_2\\
    z_3 & = x_3\\
    z_4 & = x_5\\
    z_5 & = f_w(x)=f_w(x_1,x_2,x_3,x_4,x_5,x_6)=f_w(z_1,z_2,z_3,x_4,z_4,x_6)\\
    z_6 & = f\circ F_w(x,u)=f_w(x_3,x_4,x_5,x_6,f_w(x),u)=f_w(z_3,x_4,z_4,x_6,z_5,u).
    \end{aligned}
\]

The unique solution is given by $x_1=z_1$, $x_2=z_2$, $x_3=z_3$, $x_5=z_4$, $x_6$ is the unique solution to $z_6=f_w(z_3,x_4,z_4,x_6,z_5,u)$ whose existence and uniqueness follows from the assumption, and $x_4$ is the unique solution to $z_5=f_w(z_1,z_2,z_3,x_4,z_4,x_6)$ whose existence and uniqueness also follows from the assumption.

We now use $\Phi$ as a change of coordinates to rewrite the dynamics in the coordinates $z=\Phi_u(x)$. It will suffice the consider the coordinates $z_3, \hdots, z_6$:
\[
  \begin{aligned}
  %z_1(k+1) & = & x_1(k+1)=x_3(k)=z_3(k)\notag\\
  %z_2(k+1) & = & x_2(k+1)=x_4(k)=(\Phi^{-1}(z(k)))_2\notag\\
  z_3(k+1) & = x_3(k+1)=x_5(k)=z_4(k)\\
  z_4(k+1) & = x_5(k+1)=f_w(x(k))=z_5(k)\\
  z_5(k+1) & = f_w(x(k+1))=f_w\circ F_w(x(k),u(k))=z_6(k)\\
  z_6(k+1) & = f_w\circ F_w(x(k+1),u(k+1))=f_w\circ F_w(F_w(x(k),u(k)),u(k+1)).
  \end{aligned}
\]

The last expression contains the term:
$$f_w\circ F_w(F_w(x(k),u(k)),u(k+1))=f_w(x_5(k),x_6(k),f_w(x(k)),u(k),f_w(x(k+1)),u(k+1)).$$
Given our assumption, for any $v(k)$ there exists a unique $u(k)$ so that the following equality holds:
\begin{eqnarray}
v(k)&=&f_w\circ F_w(F_w(x(k),u(k)),u(k+1))\notag\\
&=&f_w(f_w(x(k),u(k),f_w(x_3(k),x_4(k),f_w(x(k)),u(k)),u(k+1))).\notag
\end{eqnarray}

Hence, we can replace $f_w\circ F_w(F_w(x(k),u(k)),u(k+1))$ with $v(k)$ in the dynamics to obtain:
\begin{eqnarray}
%z_1(k+1) & = & z_3(k)\notag\\
%z_2(k+1) & = & g_4(z(k))\notag\\
z_3(k+1) & = & z_4(k)\notag\\
z_4(k+1) & = & z_5(k)\notag\\
z_5(k+1) & = & z_6(k)\notag\\
z_6(k+1) & = & v(k)\notag.
\end{eqnarray}
We can observe that this model is controllable since for any desired $(z_3^*,z_4^*,z_5^*,z_6^*)$, the sequence of inputs  $z_3^*, z_4^*, z_5^*,$ and $z_6^*$ will drive any state to $(z_3^*,z_4^*,z_5^*,z_6^*)$. Hence, the last $\ell=4$ tokens are controllable for this model. Noting the change of coordinates $\Phi$ uniquely determines $x_i$ for $i=3,\hdots, 6$ from $z_3,\hdots, z_6$, we conclude the last $\ell=4$ tokens are also controllable for the original dynamics. $\square$

\theendnotes

\end{document}